\documentclass[10pt,twocolumn,letterpaper]{article}
\usepackage[pagenumbers]{cvpr}
\usepackage[T1]{fontenc}
\usepackage{stfloats}
\definecolor{cvprblue}{rgb}{0.21,0.49,0.74}
\usepackage[breaklinks,colorlinks,allcolors=cvprblue,hypertexnames=true]{hyperref}
\newcommand{\shownowmathspacing}{%
  \setlength{\abovedisplayskip}{8pt plus 2pt minus 4pt}%
  \setlength{\belowdisplayskip}{8pt plus 2pt minus 4pt}%
  \setlength{\abovedisplayshortskip}{2pt plus 2pt}%
  \setlength{\belowdisplayshortskip}{5pt plus 2pt minus 3pt}%
}
\makeatletter
\g@addto@macro\normalsize{\shownowmathspacing}
\makeatother
\AtBeginDocument{\shownowmathspacing}
\newcommand{\method}{ShowNow\xspace}
\title{ShowNow: Fine-Grained Speculative Streaming for Responsive World Models}
\author{Yiren Song\qquad Guian Fang\qquad Mike Zheng Shou\textsuperscript{*}\\[3pt]
Show Lab, National University of Singapore}
\date{}
\hypersetup{pdfauthor={Yiren Song, Guian Fang, Mike Zheng Shou},pdftitle={ShowNow: Fine-Grained Speculative Streaming for Responsive World Models}}
\begin{document}
\twocolumn[{%
    \renewcommand\twocolumn[1][]{#1}%
    \maketitle
\begin{minipage}{\textwidth}
    \captionsetup{hypcap=false}
    \centering
    \includegraphics[width=\textwidth]{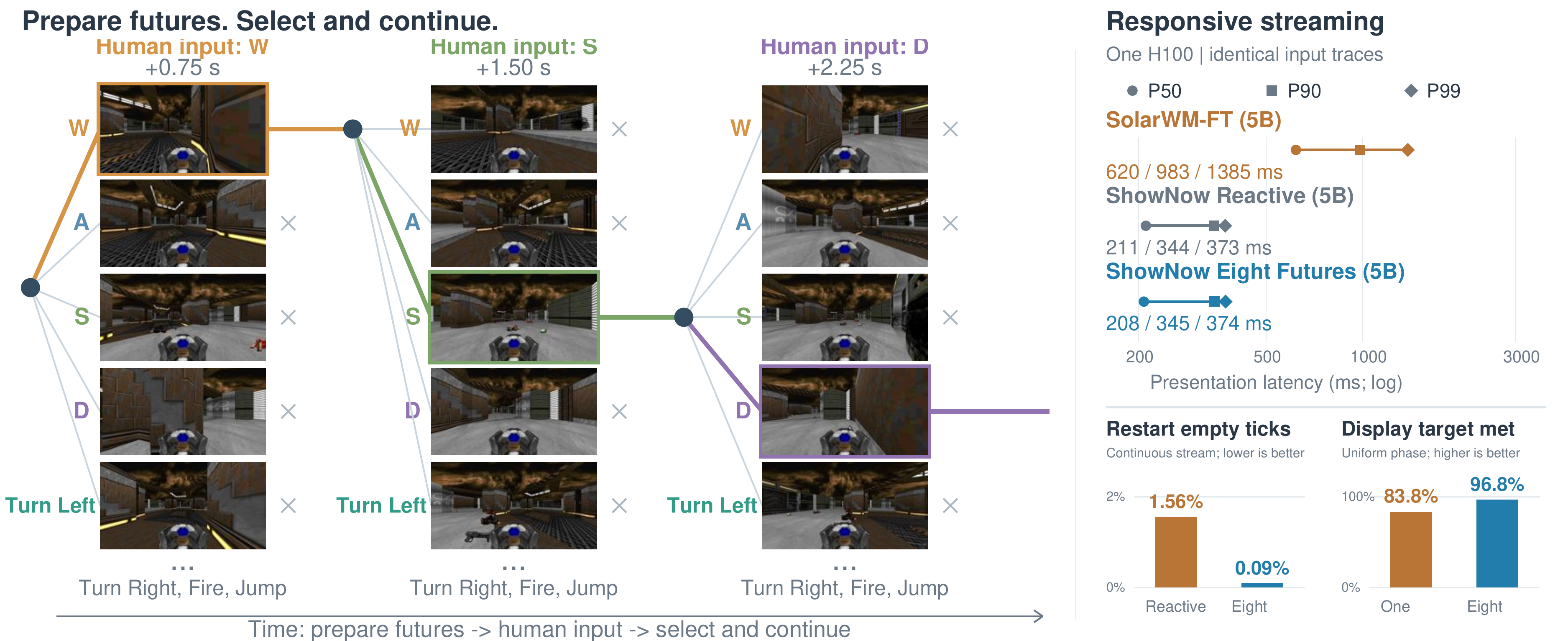}
    
    \vspace{-2.5mm}
    
    \captionof{figure}{\textbf{Prepare before input for responsive world-model streaming.} Left: matched dataset branches illustrate action-dependent outcomes. Right, top: single-H100 complete-system presentation P50/P90/P99 (log axis), including decoding and queueing. Finer commitment lowers presentation delay across trained deployments; runtime optimization enables eight-future preparation. Bottom: matched 16\,fps scheduling reduces restart empty ticks; a uniform-phase switch experiment measures the deadline benefit of action coverage.}
    \label{fig:teaser}
\end{minipage}
    \vspace{6pt}
}]
\begingroup
\renewcommand{\thefootnote}{\fnsymbol{footnote}}
\footnotetext[1]{Corresponding author.}
\endgroup

\begin{abstract}
Interactive world models must generate video continuously while responding promptly to user actions. Real-time throughput alone leaves input latency unresolved: coarse generation blocks delay action updates, and generation after input adds work to the response path. We introduce \method, a fine-grained speculative streaming framework that uses computation available during playback to prepare action-conditioned futures before input arrives. Single-latent commitment enables frequent decisions, while configurable preparation covers one to eight actions. We construct a matched multi-future dataset to train alternative continuations from a common state; shared-history tree execution and fused commitment make their preparation efficient. On a single H100, our 5B model sustains 480p generation at 16\,fps with eight prepared futures. In ViZDoom, the complete ShowNow deployment reduces median input-to-display latency from 620 to 208\,ms relative to the native-block SolarWM-FT deployment. Matched-policy experiments reduce restart empty ticks from 1.56\% to 0.09\% (94\%) and improve on-time responses to action changes. These results establish complementary roles for fine-grained commitment and advance preparation, with measured trade-offs in visual fidelity and long-horizon consistency. ShowNow turns playback-time computation into responsive visual interaction. Project: \url{https://github.com/showlab/ShowNow}.

\end{abstract}

\section{Introduction}
\label{sec:intro}
Interactive world models must sustain visual generation and respond promptly to user actions. A stream can reach its target frame rate while still displaying the consequences of an earlier command, weakening interactive control~\cite{nielsen1993response}. Occasional long delays can disrupt turns or evasive actions; in embodied and teleoperated systems, stale visual feedback can reduce the margin for correction. Responsive generation therefore requires frequent action decisions and low input-to-display latency, including its tail.

Recent models produce increasingly convincing, controllable, and streamable worlds~\cite{gamengen,matrixgame2,worldplay,solarwm}. Acceleration through fewer sampling steps and faster execution makes real-time throughput possible. Playback-paced interaction introduces another opportunity: compute can be scheduled while previously committed frames are playing. Our matched measurements reveal substantial compute slack in reactive streaming. This motivates a complementary design axis: \emph{prepare possible continuations before input arrives, then select using the latest action}. Its value depends on preparing useful candidates within the playback budget and providing frequent commitment opportunities.

\method develops this idea into fine-grained speculative streaming (Fig.~\ref{fig:teaser}). Single-latent updates shorten the commitment horizon, while configurable preparation covers one to eight possible actions. One-future preparation exploits action persistence and regenerates when the command changes~\cite{outatime,nfd}. Broader coverage prepares alternatives from the same committed state, allowing a changed action to select a ready continuation. The selected timeline becomes the common history for the next round. This couples temporal granularity, action coverage and repeated commitment in one streaming interface. Our supported actions include movement, turning, firing, and jumping.

Making this cycle practical raises two connected challenges. The model must produce distinct, action-consistent futures from a shared state, and the runtime must prepare them before their common playback deadline. We construct matched multi-future trajectories by restoring identical simulator states, supplying aligned supervision for alternative action consequences. Building on SolarWM's Wan2.2 5B backbone~\cite{solarwm}, teacher adaptation, AnyFlow distillation and Self-Forcing refinement yield a three-step causal generator~\cite{anyflow,selfforcing}. Branch-tree attention then keeps future branches isolated while reusing their historical context. Shared-prefix methods such as Hydragen~\cite{hydragen}, FlashInfer~\cite{flashinfer} and SGLang~\cite{sglang} accelerate language-model inference; we adapt this execution principle to action-conditioned multi-future video, where the user's selection determines the history carried into the next cycle. Fused commitment further moves the selected history update into generation, shortening the work remaining after a decision.

The resulting system sustains eight-future streaming at 16\,fps on one H100. A matched compute benchmark shows that shared-history execution reduces generation time by 10\% and peak memory by 48\% relative to batching eight independent sequences (Fig.~\ref{fig:response_compute}(c)). These savings make broader action coverage affordable, while the response-window analysis identifies when that coverage matters: a prepared alternative can meet a display deadline after the opportunity to regenerate has passed. This analysis links candidate readiness to interaction outcomes and motivates continuous-stream and controlled-switch evaluation.

Our experiments show how fine-grained commitment, efficient execution and advance preparation contribute to responsive streaming. Under matched runtime and read-ahead, separately trained single- and three-latent deployments demonstrate the response benefit associated with finer commitment (Table~\ref{tab:system_response}). Runtime optimization lowers commit-side work and frees preparation capacity. Within the optimized student, eight-future preparation reduces restart empty ticks by 94\% while preserving comparable presentation quantiles (Table~\ref{tab:main_latency}). Uniform-phase switches show a 13-percentage-point gain in scheduled-target attainment over One Future (Fig.~\ref{fig:response_compute}). With four actions and six sampling steps at native 16\,fps, both One Future and Four Futures reduce presentation P99 relative to Reactive (Sec.~\ref{sec:exp_ablation}). The 32\,fps study evaluates $2\times$ playback of content trained at 16\,fps and shows near-target playback with prepared futures and a shallow queue (Fig.~\ref{fig:queue_response}). These results link commitment granularity to response time and preparation to continuity and deadline readiness.

Our contributions are:
\begin{itemize}
\item \textbf{A framework for generation ahead of input.} We introduce ShowNow, which turns playback-time computation into timely action responses through fine-grained prepare--select--commit decisions. The framework unifies single- and multi-future preparation, providing a configurable compute--response trade-off for world models.
\item \textbf{A dataset and training pipeline for multiple futures.} We construct and release a matched multi-future dataset with eight action-conditioned continuations from common world states, covering held actions and action switches. Multi-future training and causal distillation equip the model to prepare alternative continuations from shared history, supporting multi-future interaction.
\item \textbf{Single-GPU real-time multi-future generation.} Shared-history tree execution and fused commitment enable a 5B model to sustain 480p generation at 16\,fps with eight prepared futures on one H100. System evaluations demonstrate substantially lower input-to-display latency for the complete deployment, while matched-policy studies establish further gains in playback continuity and on-time response through advance preparation.
\end{itemize}

\section{Related Work}
\label{sec:related}

\subsection{Interactive World Models}
GameGAN, Genie, DIAMOND, and GameNGen establish interactive visual dynamics~\cite{gamegan,genie,diamond,gamengen}. MineWorld, Matrix-Game, GameCraft, Yume, Matrix-Game 2.0, and WorldPlay extend action-conditioned streaming generation~\cite{mineworld,matrixgame,gamecraft,yume,matrixgame2,worldplay}; WorldMem, VMem, and Matrix-Game 3.5 address persistence~\cite{worldmem,vmem,matrixgame35}. Sekai and SolarWM provide data and training foundations~\cite{sekai,solarwm}. OpenWorldLib~\cite{openworldlib} organizes perception, interaction and memory within a unified world-model framework. WorldWander~\cite{worldwander} studies synchronized egocentric and exocentric video generation, broadening the viewpoints available for visual interaction. Speculative rendering in Outatime~\cite{outatime} anticipates game-engine outcomes; ShowNow instead prepares learned generative futures from a committed visual history. Next-frame speculation~\cite{nfd} motivates lightweight preparation, while ActionSplice~\cite{actionsplice} edits actions during in-flight generation. ShowNow develops configurable action coverage with matched multi-future training and shared-history execution: input selects a prepared continuation, which becomes the common history for the next round. Single-latent decisions and branch-tree execution fit this repeated prepare--select--commit cycle within a single-GPU budget.

\subsection{Controllable Video Generation}
Camera and motion controls condition video generation on geometry, trajectories and local edits~\cite{cameractrl,cameractrl2,motionctrl,trajectorycrafter}. Follow Your Pose~\cite{followpose} learns pose-guided video synthesis, Follow-Your-Click~\cite{followclick} animates selected regions with motion prompts, and MagicStick~\cite{magicstick} edits video properties through transformed control handles. Follow-Your-Creation~\cite{followcreation} uses video inpainting to support camera changes and edits in 4D creation. These controls provide complementary ways to specify visual dynamics.

Conditional video editing also develops data and architectures for preserving structure while changing appearance. VISTA~\cite{vista} learns video stylization from aligned triplets, IC-Effect~\cite{iceffect} uses contextual video conditions for effect editing, and PAI-Studio~\cite{paistudio} synthesizes motion-consistent backgrounds. PAI-Actor~\cite{paiactor} extends character replacement to multiple subjects and distills a bidirectional model into an autoregressive generator. MakeAnything~\cite{makeanything} studies coherent procedural sequences, while Mitty~\cite{mitty} and OmniHumanoid~\cite{omnihumanoid} transfer motion across human and robot embodiments. These studies demonstrate the value of aligned supervision and task-specific conditioning for controllable visual sequences. Matched action rollouts support studying divergent futures~\cite{twinrollouts}; ShowNow constructs such futures from identical ViZDoom states~\cite{vizdoom} and prepares them for repeated input-driven selection during continuous interaction.

\subsection{Efficient and Speculative Generation}
Diffusion Forcing, CausVid, Self Forcing, Self Gradient Forcing, LongLive, and Causal Forcing++ develop causal rollouts, fine-grained generation, and distillation~\cite{diffusionforcing,causvid,selfforcing,sgf,longlive,causalforcingpp}. Progressive, consistency, distribution-matching, and AnyFlow methods reduce sampling steps~\cite{progressive,consistency,lcm,dmd,dmd2,anyflow}; StreamDiffusionV2 addresses streaming schedules and rolling caches~\cite{streamdiffusionv2}. StreamingEffect~\cite{streamingeffect} applies causal distillation and online keyframe control to real-time video editing. These streaming designs motivate coordinating generation with incoming controls; ShowNow additionally prepares action alternatives before their selection deadline. SDVG accepts drafted video blocks through quality routing~\cite{sdvg}, whereas ShowNow selects futures by user action. Language-model speculative decoding accelerates generation through draft verification~\cite{specdecode,specsampling}; SpecInfer~\cite{specinfer}, Medusa~\cite{medusa} and EAGLE~\cite{eagle} develop tree-structured proposals. ShowNow applies shared-history execution to action-conditioned video futures: user input selects a branch at a playback deadline, and fused commitment installs its clean history for subsequent generation. Partitioned attention uses the standard log-sum-exp merge. FlashAttention improves the underlying attention efficiency~\cite{flashattention,flashattention2}.

\begin{figure*}[!t]
    \centering
    \includegraphics[width=\textwidth]{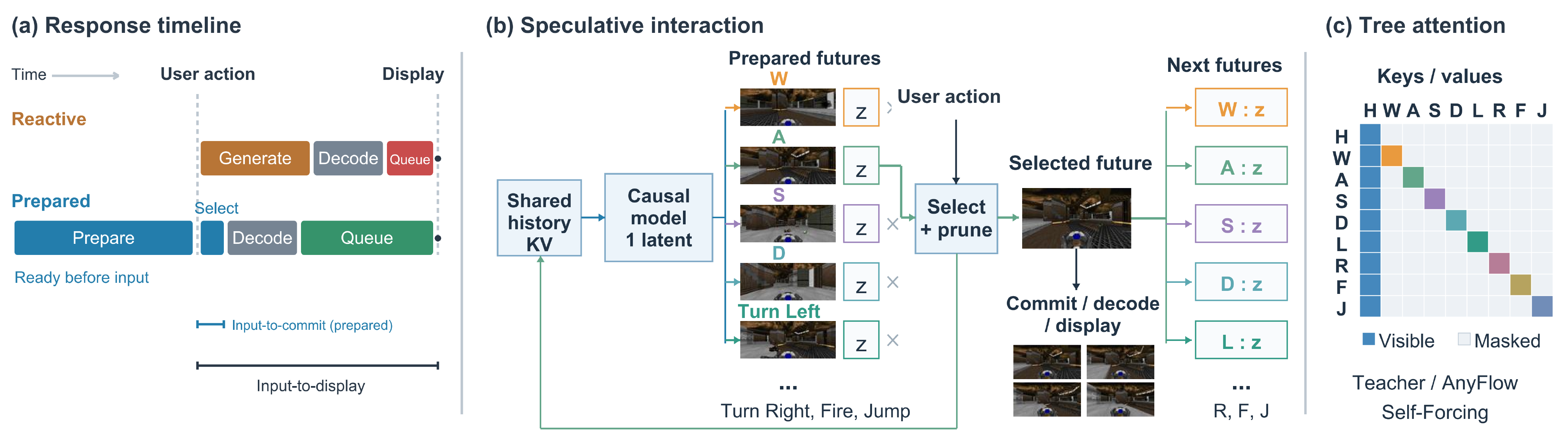}
    
\vspace{-2.5mm}
    
\caption{\textbf{Overview of \method.} (a) Preparation moves generation before the user action. Input-to-commit ends after selection; input-to-display includes decoding and queueing. Both rows share a presentation target; lengths are schematic. (b) Shared history conditions candidate futures; input selects one timeline, prunes alternatives and seeds new futures. Thumbnails illustrate dataset futures. (c) The $9\times9$ attention mask shares history H while isolating eight branches; causal students also restrict temporal visibility.}
\label{fig:architecture}
\end{figure*}

\section{Method}
\label{sec:method}

\method prepares action-conditioned futures from shared history, then selects and commits one timeline as input arrives (Fig.~\ref{fig:architecture}). The following sections describe the interaction, matched data, branch-tree model and runtime.

\subsection{Interactive Speculation}
\label{sec:setting}
Figure~\ref{fig:architecture}(a) defines the interaction task. A new command arrives while previously generated frames are playing. The system must produce its visual consequence and insert it into the continuing stream. Reactive inference reads the command before generating; speculative inference uses the preceding interval to prepare a continuation. On a candidate hit, input selects the corresponding latent immediately. The selected result is then decoded and queued for display. We measure both input-to-commit and input-to-display: the former captures model-side readiness, while the latter includes the playback schedule seen by the user. These endpoints guide generation, commitment and playback.

Let $h_n$ denote committed history and $\mathcal A=\{\mathrm W,\mathrm A,\mathrm S,\mathrm D,\mathrm L,\mathrm R,\mathrm F,\mathrm J\}$ the eight supported actions. W/A/S/D move forward, left, backward and right; L/R turn, F fires, and J jumps. At step $n$, a candidate set $\mathcal C_n\subseteq\mathcal A$ of size $K$ is prepared from the same history:
\begin{equation}
 \hat z_{n+1}^{(a)}=G_\theta(h_n,a,c_{n+1}^{(a)},\epsilon_n),\quad a\in\mathcal C_n,
\end{equation}
where $c_{n+1}^{(a)}$ is an action-derived camera condition. Input $a_n$ selects a prepared future, yielding
\begin{equation}
 h_{n+1}=\operatorname{append}(h_n,\hat z_{n+1}^{(a_n)}).
\end{equation}
Only that branch is committed and decoded. An uncovered action uses a single-candidate regeneration from the current history. The next candidate set is then prepared. Our full-coverage deployment uses $K=|\mathcal A|=8$. Restricted interfaces and partial coverage are evaluated separately.

\textbf{ShowNow, One Future} prepares the held action and regenerates immediately when the command changes, following action-persistence scheduling~\cite{nfd}. The partial-coverage heuristic at $K=1$ updates its proposal at selection points; the two policies are evaluated separately.

We denote the number of latent time steps generated per update by $B$: B1 and B3 refer to one- and three-latent blocks, respectively. This temporal block length is independent of the candidate count $K$. Temporal compression is four: one latent represents four frames, or $\Delta=250$\,ms at 16\,fps, compared with 750\,ms for three latents. Candidate preparation must fit the playback budget. We screen configurations with $Q_{0.90}(T_{\rm step})\leq\Delta$, then measure actual playback rate and empty display ticks over complete streams.

\paragraph{When coverage changes response.}
Let $L$ be the lead from a changed input to a scheduled presentation target, $C_{\rm regen}$ the cost of regenerating an uncovered action, and $C_s+D$ the selected-branch commitment and decode cost. With valid candidates ready and identical selection deadlines, the window
\begin{equation}
 C_s+D\leq L<C_{\rm regen}+C_s+D
\end{equation}
allows a prepared alternative to meet the target while regeneration cannot. Earlier inputs may let both methods meet it; later inputs may make both miss. Average gains depend on input phase, candidate availability and the playback queue.

\subsection{Multi-Future Dataset}
\label{sec:data}
We construct a matched multi-future dataset with eight action-conditioned continuations from each shared state.

Each training fork contains a common history $H$ and futures $X^{(1)},\ldots,X^{(K)}$ obtained from one saved simulator state. The ViZDoom pipeline restores that state in a fresh engine instance for each branch and applies the branch action sequence. A second replay must reproduce frames and state variables bit for bit. Fresh instances avoid rendering differences from unsaved engine state.

The eight-action collection covers W/A/S/D, left/right turning, firing and jumping. Training mixes 10,241 held-action forks with 11,471 switching forks from MAP01--28. Switching branches start with their respective actions and change keys every 0.25--1.5\,s, with 25\% held throughout. We reject damage-contaminated histories and futures. Each sample has two seconds of shared history and three seconds per future (nine history and twelve future latents). Evaluation uses separately generated states and a fixed split.

\subsection{Branch-Tree Attention}
\label{sec:tree}
We use shared-prefix tree attention to generate $K$ action-conditioned futures from one committed history. The tree enforces two structural properties: history is computed once and shared across futures, while each future remains isolated from alternative actions. This aligns computation with the prepare--select--commit cycle. Each latent slot has a lineage identifier, a temporal position, a chunk index, and a clean/noisy flag. The teacher layout permits history queries to see only history, while future queries see history and their own branch. For query $i$ and key $j$, this visibility is
\begin{equation}
M_{ij}=\begin{cases}
1,& i\in H,\ j\in H,\\
1,& i\in X^{(k)},\ j\in H\cup X^{(k)},\\
0,&\text{otherwise}.
\end{cases}
\end{equation}
Branches retain single-future temporal positions without concatenation offsets. Camera, action, and diffusion-time conditions are assigned per token.

History-only visibility decouples prefix representations from futures, so all branches reuse its computation. With matched masks and positions, tree evaluation preserves the outputs of independent branches using this history-restricted layout, up to numerical precision, removing redundant history computation.

For $N_H$ history slots and $N_F$ future slots per branch, sharing reduces the layout from $K(N_H+N_F)$ to $N_H+KN_F$ slots. With $N_H=9$, $N_F=12$, and $K=8$, this reduces the joint layout from 168 to 105 slots. Section~\ref{sec:exp_results} measures the resulting generation-time savings in streaming execution.

Causal training uses chunk-causal masks within each lineage. Action embeddings condition the transformer blocks, output head and patch tokens. Action-derived camera curves use the base model\textquotesingle s projective encoding and continue from the reached pose at each key change.

\subsection{Multi-Future Training and Distillation}
\label{sec:loss}
Training combines multi-future teacher adaptation, AnyFlow causal distillation~\cite{anyflow,solarwm}, and Self-Forcing refinement~\cite{selfforcing}. The teacher adapts the SolarWM/Wan2.2 backbone with bidirectional branch-tree attention and branchwise flow matching. Matched branches receive common clean history and shared noise, with action-specific camera conditions. Teacher and initial student training retain a branch-difference term ($\lambda=0.25$), with mixed cameras for deployment alignment.

For clean future $x_0^{(k)}$ in branch $k$, we draw shared noise $\epsilon$ and noise level $\sigma$, forming $x_\sigma^{(k)}=(1-\sigma)x_0^{(k)}+\sigma\epsilon$. The target velocity is $u^{(k)}=\epsilon-x_0^{(k)}$, and branchwise flow matching minimizes
\begin{equation}
\label{eq:flow_matching}
\mathcal L_{\mathrm{FM}}=\frac{w(\sigma)}{K}\sum_{k=1}^{K}\|v_\theta^{(k)}-u^{(k)}\|_2^2,
\end{equation}
where $v_\theta^{(k)}$ is the predicted velocity, $w(\sigma)$ weights the noise level, and the norm averages over future elements. Shared history and noise align the branches while their action conditions determine distinct continuations.

AnyFlow transfers the teacher to a few-step student using tree teacher forcing. Chunk-causal attention uses one latent per chunk and up to fifteen prior chunks. Self-Forcing then refines the student on its own autoregressive histories using self-gradient forcing and distribution matching~\cite{sgf,dmd}. The final eight-action student uses three denoising steps; validation action accuracy selects the checkpoint, with LPIPS breaking ties.

\subsection{Efficient Streaming Inference}
\label{sec:runtime}
\paragraph{Shared-prefix cache.}
FastTree keeps raw, pre-positional keys and values for the most recent fifteen committed latents in a ring. As the window slides, it reapplies window-relative rotary and projective positional transformations to that prefix. This avoids recomputing committed latents.

\paragraph{Partitioned attention.}
Each candidate attends to shared-history and branch-local partitions. Let $o_H,o_a$ denote their normalized outputs and $\ell_H,\ell_a$ their log normalizers. We combine the partitions as
\begin{equation}
 o=\frac{e^{\ell_H}o_H+e^{\ell_a}o_a}{e^{\ell_H}+e^{\ell_a}},
\end{equation}
using a stable log-sum-exp merge. This recovers attention over the shared prefix and the selected branch up to numerical precision. The history cache is stored once, and each branch keeps its own future keys and values, reducing duplicated memory traffic as the candidate count grows.

\paragraph{Candidate preparation and commitment.}
Static shapes enable CUDA graphs; positional transforms are fused and transformer blocks compiled. The deployment prepares eight candidates. The selected branch supplies the committed history; in the optimized path, its clean-time K/V update is fused into the next generation pass, so decoding can begin immediately after selection. A small streaming TAEHV decoder produces its frames.

\paragraph{Playback scheduling.}
Candidate generation overlaps display of the previous block. The demo reads the held key when a short playback lead remains, then commits, decodes, and presents the selected future. We separately measure generation, commitment, decoding, and presentation.

We compare streaming and cache-free inference with matched actions and cameras in Sections~\ref{sec:exp_results} and \ref{sec:exp_ablation}.

\section{Experiments}
\label{sec:experiments}
We evaluate continuous response and compute (E1), short-horizon control and quality (E2), and one-minute generation (E3). The main experiments use GPU measurements. Appendices~\ref{sec:supp_protocol}--\ref{sec:supp_test} provide training configurations, full statistical protocols and extended ablations.

\subsection{Settings and Measurement}
\label{sec:exp_setup}
\paragraph{Model and controls.}
SolarWM-FT retains the native three-latent block after fine-tuning on the same data with the same teacher and post-training stages. Timing and quality evaluation use the same trained SolarWM-FT model, with FastTree execution and automatic four-frame lead. The ShowNow student uses three sampling steps, one latent per update, 15 history latents, and optimized BF16/FA3 execution with fused commitment. Within ShowNow, Reactive generates after input, One Future prepares the held action and regenerates on a change, and multi-future configurations cover supported actions. Matched comparisons share checkpoint, runtime, traces and TAEHV. A queue threshold of one or zero remaining frames triggers selection; each update appends four frames without flushing.

\paragraph{Continuous streams and endpoints.}
The main continuous-stream experiments use six seeds and four refractory-Poisson input rates (0.5/1/2/4 events per second), totaling 1857 scored presses per policy. Each trace lasts 60\,s: inputs in the first 50\,s are scored, with the final 10\,s reserved for observing responses. We report P50/P90/P99 of the equal-rate mixture $F_{\rm mix}(t)=\frac14\sum_r F_r(t)$. Presentation latency includes commitment, decoding and software queueing; primary latency ends at latent commitment. Presentation marks the first displayed command-conditioned frame; action-effect detection additionally assesses visible control response. Matched ShowNow policies share checkpoints, runtime and input traces. Table~\ref{tab:system_response} compares complete trained deployments, including their block sizes and runtimes. Every scored input is retained: superseded commands wait for the next served command. Confidence intervals resample paired seed clusters. Empty ticks count missing display frames during held input. Restart ticks precede the first response after idle; ongoing ticks follow it. Both are normalized by held-key ticks, excluding initial startup.

\paragraph{Controlled response window.}
A separate study executes 2650 matched GPU pairs across ten scenes and six H100s. Each pair shares history, input, GPU, a 6\,ms selection guard, four-frame FIFO and presentation target, with balanced method order. The final changed action is held through its response. The study includes a complete lead sweep and 500 uniform-phase pairs per interface. Target attainment is the fraction of trials whose first new-action frame is scheduled by the common display deadline, with a 2\,ms tolerance.

\paragraph{Control and quality protocols.}
E2 uses 48 states and eight three-second futures; E3 uses ten one-minute scenes with changes every 2\,s after 15 real introductory latents. Independent testing uses 96 E2 states and 20 E3 scenes. The frozen switch-aware IDM and latent feature distance operate on latents, independently of decoding. E2 LPIPS, PSNR and SSIM use Wan VAE; runtime presentation uses TAEHV. All models share masks and scoring rules. SolarWM-FT quality and timing use the same B3 model.

\begin{table*}[!tb]
\caption{\textbf{Trained-model deployment response on one H100.} Input-to-display quantiles in ms, using the six-seed equal-rate protocol (1857 inputs). SolarWM-FT and ShowNow use separately trained B3 and B1 models. FastTree uses FA2; optimized B1 uses FA3 in its deployment environment, with the B1 checkpoint fixed across runtimes. Optimized rows use one-frame lead; SolarWM-FT automatic lead is four frames. Empty ticks are normalized by held-key ticks.}
\centering\small
\setlength{\tabcolsep}{4pt}
\begin{tabular*}{\textwidth}{@{\extracolsep{\fill}}lrrrrr@{}}
\toprule
Deployment & P50 & P90 & P99 & FPS & Restart / ongoing empty (\%)\\
\midrule
SolarWM-FT B3, FastTree (automatic lead) & 620 & 983 & 1385 & 15.76 & 1.49 / 0.00\\
SolarWM-FT B3, FastTree (one-frame lead) & 536 & 915 & 1370 & 14.50 & 3.48 / 5.91\\
ShowNow B1, FastTree (one-frame lead) & 210 & 339 & 374 & 15.61 & 2.43 / 0.00\\
ShowNow Reactive, optimized & 211 & 344 & 373 & 15.75 & 1.56 / 0.00\\
ShowNow Eight Futures, optimized & 208 & 345 & 374 & 15.99 & 0.09 / 0.00\\
\bottomrule
\end{tabular*}

\label{tab:system_response}
\end{table*}

\begin{figure*}[!tb]
\centering
\includegraphics[width=\textwidth]{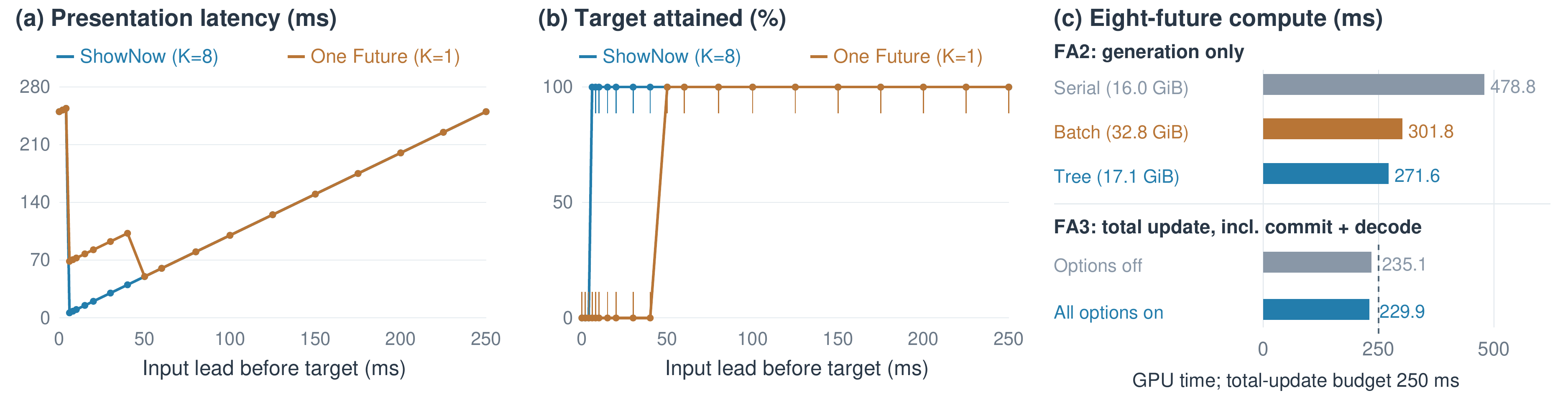}

\vspace{-2.5mm}

\caption{\textbf{When prepared alternatives improve response.} GPU measurements. (a) Median action-to-presentation latency versus input lead before a fixed target, for eight supported actions at 16\,fps. (b) Fraction reaching that target within 2\,ms. Each point uses 30 matched pairs; bars show 95\% median-bootstrap or Wilson intervals. The 6--40\,ms window exposes regeneration delay; the full phase sweep includes inputs too late for either method and early enough for both. (c) Matched GPU benchmarks: upper rows compare generation in FA2 (parentheses give peak allocated memory); lower rows compare total generation, commitment and decoding in FA3. Each uses 100 updates after ten warm-ups on one H100. The 250\,ms line applies to total updates; host scheduling and display waits are excluded.}
\label{fig:response_compute}
\end{figure*}

\subsection{Response and Preparation Cost}
\label{sec:exp_results}
\paragraph{Continuous presentation.}
SolarWM-FT records 620/983/1385\,ms presentation P50/P90/P99 with automatic lead. Matching one-frame lead and FastTree execution across the trained B3 and B1 deployments gives medians of 536 and 210\,ms, respectively (Table~\ref{tab:system_response}). B3's shorter lead trades response against playback continuity, while optimized B1 execution frees compute for preparation. The eight-future deployment achieves 208/345/374\,ms at 15.99\,fps. At zero-frame threshold, matched Reactive/One/Eight presentation triples are 152/279/392, 148/279/312 and 144/277/312\,ms (Table~\ref{tab:main_latency}).

\begin{table*}[!tb]
\caption{\textbf{Eight-action control and paired quality.} 48 validation states used for checkpoint selection, eight futures each. Shared frozen IDM, engine-state mask and Wan VAE. Teacher evaluation uses 30 denoising steps; ShowNow uses the real-time path. Transition spans two updates after switching; delay is relative to real video in 0.25\,s latents.}
\centering\small
\setlength{\tabcolsep}{4pt}
\resizebox{\textwidth}{!}{%
\begin{tabular}{lcccccc}
\toprule
Model / inference & Action acc.$\uparrow$ & Transition/settled$\uparrow$ & Delay$\downarrow$/not slower$\uparrow$ & LPIPS 0--0.5/0--1\,s$\downarrow$ & PSNR 1\,s$\uparrow$ & SSIM 1\,s$\uparrow$ \\
\midrule
Teacher & \textbf{.943} & \textbf{.936}/\textcolor{cvprblue}{.949} & \textbf{.00}/\textbf{.97} & \textbf{.344}/\textbf{.380} & \textbf{19.61} & \textbf{.574} \\
SolarWM-FT & .793 & .591/\textbf{.969} & +.44/.68 & \textcolor{cvprblue}{.355}/\textcolor{cvprblue}{.401} & \textcolor{cvprblue}{18.80} & \textcolor{cvprblue}{.560} \\
ShowNow & \textcolor{cvprblue}{.860} & \textcolor{cvprblue}{.813}/.901 & \textcolor{cvprblue}{+.20}/\textcolor{cvprblue}{.82} & .399/.437 & 17.64 & .530 \\
\bottomrule
\end{tabular}}

\label{tab:main_quality}
\end{table*}

\paragraph{Using the playback-time compute budget.}
In matched six-seed 16\,fps runs, Reactive/One Future use 19\%/25\% GPU busy time. Eight futures use 80\%, sustain 15.99\,fps and reduce restart empty ticks from 1.56\% to 0.09\%, with comparable presentation quantiles (208/345/374 versus 211/344/373\,ms; Table~\ref{tab:main_latency}). GPU busy fraction measures generation, commitment and decoding GPU time divided by wall time. This operating point converts playback-time compute capacity into broader action coverage while preserving real-time playback.

\begin{table}[!tb]
\caption{\textbf{Preparation and playback scheduling.} ShowNow policies share checkpoint, runtime and input traces: six seeds, 1857 scored presses each (first 50 s of 60 s traces), equal-rate-mixture display quantiles in ms. Lead specifies the pending-frame threshold for selection. The 32 FPS setting evaluates $2\times$ playback of content trained at 16 fps; FPS reports achieved playback throughput.}
\centering\fontsize{8}{9.5}\selectfont
\setlength{\tabcolsep}{2pt}
\begin{tabular*}{\columnwidth}{@{\extracolsep{\fill}}lrcrr@{}}
\toprule
Policy & Lead & P50/P90/P99 & FPS & \shortstack{Empty (\%)\\restart/ongoing}\\
\midrule
\multicolumn{5}{l}{\textit{8 actions, 16 fps}}\\
Reactive & 1 & 211/344/373 & 15.75 & 1.56/0.000\\
One Future & 1 & 211/344/374 & 15.79 & 1.28/0.000\\
Eight Futures & 1 & 208/345/374 & 15.99 & 0.09/0.000\\
Reactive & 0 & 152/279/392 & 15.71 & 1.82/0.000\\
One Future & 0 & 148/279/312 & 15.77 & 1.47/0.000\\
Eight Futures & 0 & 144/277/312 & 15.99 & 0.09/0.000\\
\midrule
\multicolumn{5}{l}{\textit{4 actions, 32 fps playback ($2\times$ speed)}}\\
Reactive & 1 & 106/169/184 & 31.38 & 1.92/0.008\\
One Future & 1 & 106/169/185 & 31.52 & 1.49/0.003\\
Four Futures & 1 & 83/148/176 & 31.95 & 0.14/0.005\\
Reactive & 0 & 119/199/216 & 25.26 & 5.20/15.875\\
One Future & 0 & 78/141/154 & 31.34 & 1.51/0.554\\
Four Futures & 0 & 73/137/153 & 31.96 & 0.14/0.000\\
\bottomrule
\end{tabular*}

\label{tab:main_latency}
\end{table}

\paragraph{Readiness, coverage and presentation.}
Preparation reduces median input-to-commit from 136\,ms for Reactive to 86/88\,ms for one/eight futures. Queueing can align prepared and regenerated responses to the same display slot. Preparation becomes valuable near deadlines or with shallow queues, where regeneration can delay display. All three policies have no observed ongoing empty ticks at the one-frame threshold (Table~\ref{tab:main_latency}).

Figure~\ref{fig:response_compute}(a,b) shows how preparation covers action changes close to a presentation deadline. At tested leads from 6 to 40\,ms, eight prepared alternatives meet the shared target in every measured trial, whereas one-future preparation requires regeneration and waits approximately one additional frame. Across all 500 uniform-phase pairs, coverage increases target attainment by 13.0 percentage points, from 83.8\% to 96.8\%, with a mean presentation gain of 9.2\,ms (paired 95\% CI [7.0,12.0]). This identifies the mechanism's practical benefit: keeping the next display deadline reachable when input changes late.

\begin{figure}[!tb]
\centering
\includegraphics[width=\columnwidth]{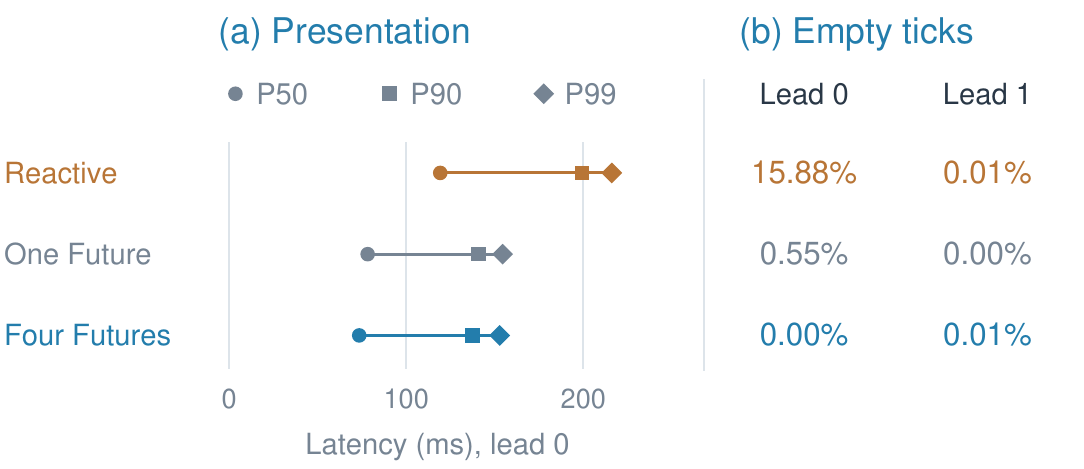}

\vspace{-2.5mm}

\caption{\textbf{Preparation with shallow queues.} Six-seed matched GPU runs, four actions, 32\,fps playback ($2\times$ the training rate). (a) Input-to-display P50/P90/P99 at zero-frame threshold. (b) Ongoing empty ticks at zero-/one-frame thresholds. Lower is better; thresholds control selection timing.}
\label{fig:queue_response}
\end{figure}

\paragraph{Tighter playback budgets.}
In a 32\,fps accelerated-playback stress test ($2\times$ playback of 16\,fps-trained content), preparation allows a shallow queue while sustaining near-target throughput (Fig.~\ref{fig:queue_response}). At zero-frame threshold, Reactive reaches 25.26\,fps with 15.88\% ongoing empty ticks; One Future reaches 31.34\,fps, and Four Futures reaches 31.96\,fps with no observed ongoing empty ticks. The optimized Reactive path starts when no undisplayed frame remains, exposing about 50.1\,ms of generation with fused commitment plus 3.5\,ms decoding. A display tick is 31.25\,ms; prepared candidates remove generation from this empty-queue response path. Tick losses depend on scheduling phase as well as mean update cost.

Raising Reactive's threshold to one frame restores 31.38\,fps and reduces ongoing empty ticks to 0.008\%, with presentation P50/P90/P99 of 106/169/184\,ms. Four Futures at threshold zero gives 73/137/153\,ms, illustrating the measured buffering--latency trade-off. Table~\ref{tab:main_latency} reports the full policy--threshold comparison on shared input traces. Quality in Table~\ref{tab:main_quality} is evaluated at the 16\,fps training rate. At threshold zero, Four versus One Future further reduces P50/P90/P99 by 5/4/2\,ms; the P90/P99 confidence intervals exclude zero. Thus preparation and buffering jointly set the response budget, with extra coverage improving readiness and continuity.

\begin{figure*}[!tb]
\centering
\begin{minipage}{0.49\textwidth}
\includegraphics[page=1,width=\linewidth]{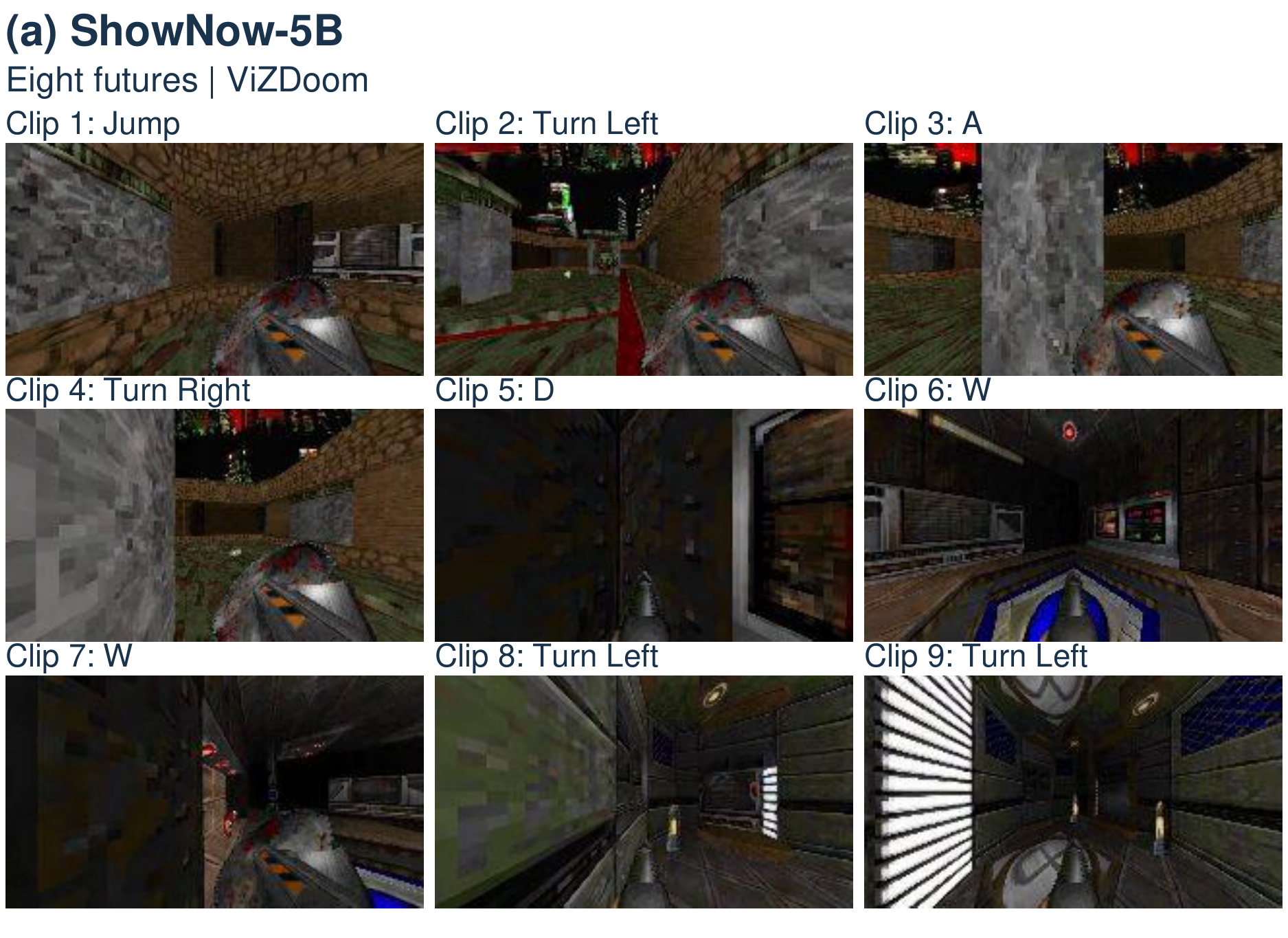}
\end{minipage}\hfill%
\begin{minipage}{0.49\textwidth}
\includegraphics[page=1,width=\linewidth]{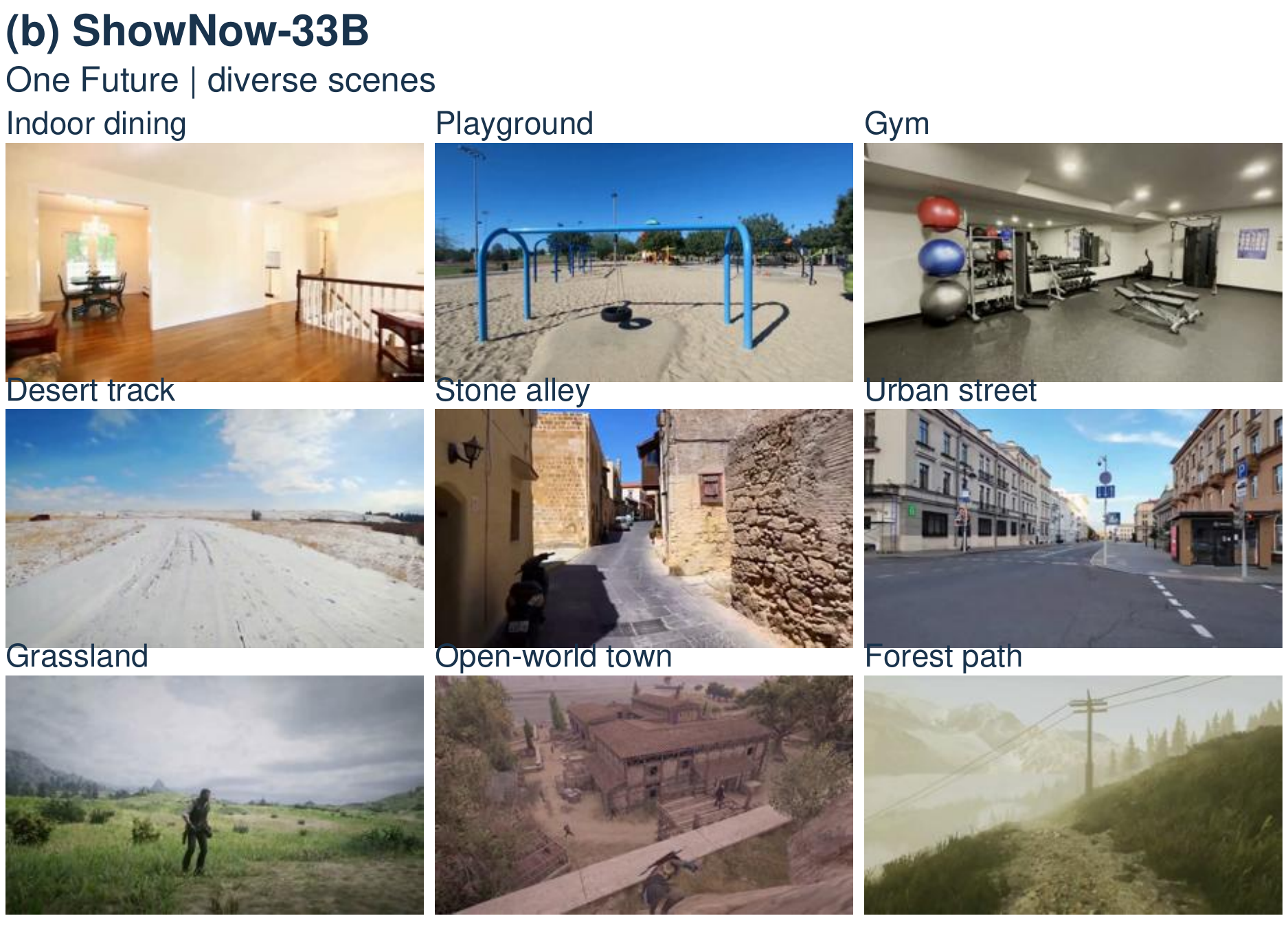}
\end{minipage}

\vspace{-2.5mm}

\caption{\textbf{ShowNow at two model scales.} (a) Nine eight-future 5B ViZDoom clips, labeled by action. (b) Nine single-latent ShowNow-33B sequences across indoor, outdoor and game scenes, following recorded camera trajectories at $1344\times768$. Panels show the first frame of each sequence; videos are provided as ancillary files.}
\label{fig:qualitative_results}
\end{figure*}

\paragraph{Compute and candidate coverage.}
Shared-history execution reduces eight-candidate generation from 301.8 to 271.6\,ms against batch-eight independent sequences in a matched FA2 benchmark, with peak allocated memory falling from 32.8 to 17.1\,GiB. Serial independent generation takes 478.8\,ms. In a separate matched FA3 benchmark, the tree fits the 250\,ms GPU update budget with all deployment optimizations disabled (235.1\,ms) and enabled (229.9\,ms), including commitment and decoding. Fused commitment shortens the post-decision path from approximately 17.5 to 3.5\,ms. Figure~\ref{fig:response_compute}(c) separates the shared-history and runtime comparisons.

\subsection{Control, Quality and Ablations}
\label{sec:exp_ablation}
Single-latent commitment performs 240 autoregressive updates per minute, compared with 80 for B3, enabling more frequent action decisions. In the first 0.5\,s after a switch, accuracy is 0.813 versus 0.591 for SolarWM-FT; settled accuracy is 0.901 versus 0.969 (Table~\ref{tab:main_quality}). Overall accuracy is 0.860, while validation LPIPS is 0.437 versus normal inference's 0.416. On 500 isolated switches, IDM-detected action-effect median/P90 latency falls from SolarWM-FT's 603/987\,ms to ShowNow's 297/815\,ms. Figure~\ref{fig:qualitative_results}(a) shows generated ViZDoom clips under changing actions.

Across nine paired noise seeds on 20 held-out one-minute scenes, the deployed path has mean FD 11.4 and key accuracy .772; disabling deployment fusions gives 8.3 and .785. The FD difference is sensitive to seed and scene variation. Three-seed normal inference and SolarWM-FT references give FD 6.3/3.4 and key accuracy .801/.912. At 80 aligned autoregressive steps, their FD is 4.0/4.1 and deployed B1 is 9.8. These results show that both deployment execution and longer autoregressive histories affect rollout quality.

\paragraph{Generation cost and tail response.}
In the native 16\,fps four-action sweep, six-step sampling costs 89\,ms per candidate. Preparation reduces Reactive's presentation P99 from 498 to 373--374\,ms: One Future gives a paired difference of $-124$\,ms (95\% CI $[-188,-100]$), and Four Futures gives $-125$\,ms ($[-189,-103]$). Four Futures also reduces restart empty ticks from Reactive's 2.77\% to 0.03\%.

\paragraph{ShowNow-33B.}
We extend ShowNow to a 33B camera-conditioned backbone by fine-tuning SolarWM-H3-33B for single-latent generation with One Future scheduling. At $1344\times768$ resolution, generation and history commitment use eight-way sequence parallelism on B300 GPUs. With measured GPU generation and 47\,ms per-latent decoding modeled on a dedicated ninth GPU, ShowNow-33B reduces median input-to-display latency from 862 to 232\,ms (73\%) and P99 from 1202 to 330\,ms (73\%) relative to the native deployment, while achieving 23.94\,fps against a 24\,fps target. These results demonstrate the framework's applicability to larger models and high-resolution streaming. Figure~\ref{fig:qualitative_results}(b) shows nine generated sequences spanning indoor, outdoor and game scenes under recorded camera trajectories. Appendices~\ref{sec:h3_extension} and~\ref{sec:supp_qualitative} contain the full 33B timing setup and expanded visual sequences.

\section{Conclusion}
We presented \method, a streaming framework combining single-latent commitment, shared-history execution and advance preparation of action-conditioned futures. The system supports eight prepared futures at 16\,fps on one H100. Matched-runtime comparisons link finer commitment to lower presentation latency. Execution optimizations reduce candidate cost and expand preparation capacity. Within the same student, preparation improves restart continuity and scheduled-target attainment while maintaining comparable presentation quantiles at the default operating point. At higher sampling cost, both single- and multi-future preparation improve presentation P99. Controlled switches and queue sweeps relate these benefits to candidate readiness, input timing and playback budgets, including shallow-buffer accelerated playback. Fine-grained autoregression also brings fidelity and long-horizon consistency trade-offs.

\clearpage
\appendix
\setcounter{table}{0}
\setcounter{figure}{0}
\setcounter{equation}{0}
\renewcommand{\thetable}{S\arabic{table}}
\renewcommand{\thefigure}{S\arabic{figure}}
\renewcommand{\theequation}{S\arabic{equation}}
\renewcommand{\theHtable}{appendix.\arabic{table}}
\renewcommand{\theHfigure}{appendix.\arabic{figure}}
\renewcommand{\theHequation}{appendix.\arabic{equation}}
\section{Evaluation Protocols}
\label{sec:supp_protocol}
This appendix provides the protocols, implementation details and extended measurements supporting the main paper. Sections~\ref{sec:supp_training_objectives}--\ref{sec:supp_compute} describe training and execution; Sections~\ref{sec:stream_details}--\ref{sec:h3_extension} expand the response and quality studies. Section~\ref{sec:supp_qualitative} presents dataset examples and generated sequences. Figures and tables in the appendix carry an S prefix; citations share the bibliography with the main text.

\paragraph{Evaluation sets and configurations.}
E1 measures continuous response, E2 measures short-horizon control and paired quality, and E3 measures one-minute rollouts. E1 uses six seeds, four input rates and 60\,s per trace, with 1857 scored presses per policy. The main interface has eight actions at 16\,fps; the accelerated-playback interface restricts the same model to WASD at 32\,fps, playing 16\,fps-trained content at $2\times$ speed. E2 validation uses 48 states with eight three-second futures; independent testing uses 96 states. E3 validation uses ten scenes and independent testing uses 20 scenes. B1/B3 denote one-/three-latent update lengths; $K$ denotes the number of candidate futures. One latent represents four video frames. Multi-future response comparisons use full coverage: eight of eight actions or four of four. One Future denotes held-action preparation with immediate regeneration on a key change.

\paragraph{Input traces and response accounting.}
Input events begin at 0.5\,s and use gaps $0.15+\mathrm{Exp}(1/\lambda-0.15)$ seconds, where the exponential argument denotes its mean and $\lambda\in\{0.5,1,2,4\}$ events/s. While a key is held, an event releases it with probability 0.2 and otherwise switches to a different key. An event during idle presses a key. New keys are sampled among the least-used eligible actions, with random tie-breaking; a key remains held until changed or released. Six trace seeds (0--5) give 120/231/505/1001 scored presses at the four respective event rates. Inputs in the first 50\,s are scored; the final 10\,s observe their responses. A direct response matches the requested action before it is superseded. Superseded inputs wait for the next served command, including release-to-repress intervals. An unobserved successor receives infinite latency; every scored input in the six-seed runs has an observed response.

The equal-rate-mixture CDF is $F_{\rm mix}(t)=\frac14\sum_r F_r(t)$, where $F_r$ is the empirical CDF within rate $r$. Quantiles use the generalized inverse $\inf\{t:F_{\rm mix}(t)\geq p\}$, equivalently assigning weight $1/(4N_r)$ to each event at rate $r$. Paired confidence intervals use 2000 seed-cluster bootstrap draws, retaining each seed's four rates and all matched policies. Table~\ref{tab:rate_detail} gives the per-rate results for the main deployments.

\paragraph{Display and compute endpoints.}
Input-to-commit ends when the selected latent is available to decode. Input-to-display includes decoding and software queueing; network and physical-display delays are excluded. Display ticks occur at the requested playback rate. A restart empty tick precedes the first response after idle; an ongoing empty tick follows it during held input. Both are normalized by held-key ticks, excluding initial startup. The queue threshold specifies the number of pending frames that triggers a decision; each committed latent appends four frames without flushing. GPU busy fraction divides recorded generation, commitment and decoding time by full-trace wall time.

\paragraph{E2 control and fidelity.}
The frozen switch-aware IDM reads three consecutive Wan latents. A shared engine-state mask confirms motion along the requested translation axis, yaw changes above $5^\circ$, visible attack state, or jump take-off. The same real-branch mask applies to every model, covering 76.8\%/77.8\% of validation/test latents. Transition uses latent ages 0 and 1 within each action segment (0.5\,s at 16\,fps); settled accuracy uses ages $\geq2$. The same latent-index rule and real-branch mask apply to B1 and B3. Response delay compares the first generated and real recognition of the same switch, conditional on detection in both. LPIPS, PSNR and SSIM use matched real futures and Wan-VAE decoding. The teacher uses 30 denoising steps; deployed students use three.

The scorer is trained on held-action and switching clips, split by source trajectory and excluding evaluation-fork starts. It reaches 0.986/0.985 accuracy on real E2 validation/test latents. On all validation latents without masking, real/teacher/SolarWM-FT/normal/ShowNow accuracy is .951/.909/.808/.844/.818. On 1808 real switches from 80 held-out-map forks, it detects 1740 (96.2\%) within the new segment, including 1649 (91.2\%) at the switch latent. These calibration clips are separate from the generated-video test states.

\paragraph{E3 rollout quality.}
After 15 real introductory latents, each scene generates 60\,s with a new key every 2\,s; forward motion is twice as likely. Normal inference replays ShowNow's recorded keys and nominal cameras. Key accuracy scores settled latents, at least two latent time steps (0.5\,s) after switching. Feature distance (FD) compares 96-dimensional latent channel mean/std features against real video from the same scenes. Features are pooled across scenes within each five-second window; minute-average FD averages twelve windows. E3 uses TAEHV for display, while latent metrics are decoder-independent. Multi-seed results and their uncertainty appear in Sec.~\ref{sec:supp_test}.

\section{Training and Implementation}
\label{sec:supp_training_objectives}

Training has three stages: multi-future adaptation, AnyFlow distillation~\cite{anyflow,solarwm}, and Self-Forcing refinement~\cite{selfforcing}.

\paragraph{Stage 1: Multi-future teacher training.}
The teacher descends from the SolarWM/Wan2.2 backbone~\cite{solarwm} and is adapted with bidirectional branch-tree attention. Each matched branch receives a common clean history and its own action and camera conditions. For branch $k$, $x_\sigma^{(k)}=(1-\sigma)x_0^{(k)}+\sigma\epsilon$ and $u^{(k)}=\epsilon-x_0^{(k)}$. Branchwise flow matching follows Eq.~\ref{eq:flow_matching}, with the backbone scheduler's timestep weight $w(\sigma)$ and mean squared error over future elements. For $K>1$, the shared-noise branch-difference auxiliary term is
\begin{equation}
\mathcal L_{\mathrm{diff}}=\frac{2w(\sigma)}{K(K-1)}
\sum_{a<b}\left\|(v_\theta^{(a)}-v_\theta^{(b)})-(u^{(a)}-u^{(b)})\right\|_2^2.
\end{equation}
Shared noise gives $u^{(a)}-u^{(b)}=x_0^{(b)}-x_0^{(a)}$, so this term supervises differences between action-conditioned futures. Teacher training uses $\mathcal L_{\mathrm{FM}}+0.25\mathcal L_{\mathrm{diff}}$; the auxiliary term is zero for $K=1$. Switching adaptation continues from the held-action teacher.

Training mixes real, nominal and static camera conditions as described under Condition alignment below.

\paragraph{Stage 2: AnyFlow causal distillation.}
AnyFlow flow-map distillation transfers the teacher to a few-step student using tree teacher forcing~\cite{anyflow,solarwm}. The ShowNow student uses one-latent causal chunks with up to fifteen history latents; the SolarWM-FT student uses three-latent chunks. Initial student training also applies $0.25\mathcal L_{\mathrm{diff}}$ on auxiliary updates, evaluating the instantaneous velocity at equal start and end times in the AnyFlow parameterization.

\paragraph{Stage 3: Self-Forcing refinement.}
We refine the student on its own autoregressive histories through Self-Forcing~\cite{selfforcing}, using self-gradient forcing and distribution matching~\cite{sgf,dmd}. Cache-free tree rollouts are replayed with teacher forcing to supply gradients; the adapted teacher and learned critic score samples using bidirectional tree attention. The critic updates each iteration and the student every fifth iteration. Unconditional teacher guidance removes both action and future camera motion. The final eight-action student uses three denoising steps.

\label{sec:supp_implementation}
\paragraph{Training configurations.}
\label{sec:training_configuration}
SolarWM-FT and ShowNow use the same adapted teacher, training data and AnyFlow/distillation stages, with three- and single-latent blocks, respectively. Their student action learning rates are $10^{-4}$ and $3\times10^{-4}$, respectively; the initial student branch-difference loss weights are 0 and 0.25. Timing and quality use the same trained model within each deployment; the B1 student is fixed across scheduling policies.

\paragraph{Branch conditioning.}
Our generator uses the SolarWM/Wan2.2 5B latent video transformer backbone~\cite{solarwm}. Action embeddings modulate transformer blocks and the output head through timestep conditioning. We also add a zero-initialized action projection to patch-embedded tokens, providing a direct residual signal.

Camera control uses the base model's projective positional encoding. History uses recorded poses. At deployment, action-specific nominal curves specify translation and yaw, including view rotation for turning and vertical motion for jumping. At a key change, the next curve starts from the reached pose, so translation follows the updated view. Firing is represented through the action condition. Camera coordinates are rebased to the first latent and follow the base model's frame-alignment convention. Simulator translation is scaled by 0.5; intrinsics use the base convention.

\paragraph{Condition alignment.}
To accommodate collisions and deployment conditions, training mixes real, nominal, and static future cameras. The final eight-action Stage~3 configuration uses real-camera probability 0.3 and static-camera probability 0.1. Nominal paths revert to real cameras when actual travel is below half the nominal distance. An unconditional mode removes future action and freezes the camera; single-trajectory examples use real or static/unconditional conditions. Evaluation uses nominal cameras.

\paragraph{Causal training masks.}
For causal teacher forcing, clean queries see clean tokens from their lineage up to their own chunk. Noisy queries see prior clean chunks within a bounded window and noisy tokens in their own chunk. A cache-free autoregressive window uses the corresponding chunk-causal mask for a clean prefix and candidate noisy blocks. Training uses compiled block masks.

\section{Runtime and Preparation Cost}
\label{sec:supp_compute}
Figure~\ref{fig:response_compute}(c) separates shared-history savings from deployment optimizations. Both benchmarks use the B1 student, three sampling steps, 15 history latents, BF16, $864\times480$ resolution and an exclusive H100. Each benchmark uses ten warm-up updates followed by 100 CUDA-event measurements.

\paragraph{Shared history.}
Each shared-history benchmark configuration is built and freed separately. The independent batch runs eight sequences with duplicated persistent history K/V caches. Under FA2, shared-tree execution reduces generation from 301.8 to 271.6\,ms and peak allocated memory from 32.8 to 17.1\,GiB (Table~\ref{tab:matched_tree_compute}). Serial independent generation takes 478.8\,ms. The benchmark uses one scene. With identical history, keys, cameras and noise, tree/batched three-step outputs have mean absolute differences of .022/.027 and relative $L_2$ differences of 3.7\%/4.7\% against serial BF16 execution. This check compares candidates from one shared history; Sec.~\ref{sec:supp_test} evaluates accumulated rollout differences.

\begin{table}[!tb]
\caption{\textbf{Matched eight-future generation.} CUDA-event times in ms; peak allocated memory includes each configuration's buffers and graph pool. These are generation-only measurements.}
\centering\small
\setlength{\tabcolsep}{3pt}
\begin{tabular}{lrrr}
\toprule
FA2 generation & P50 & P99 & GiB\\
\midrule
Serial independent & 478.8 & 481.0 & 16.0\\
Batched independent & 301.8 & 304.3 & 32.8\\
Shared tree & 271.6 & 274.3 & 17.1\\
\bottomrule
\end{tabular}

\label{tab:matched_tree_compute}
\end{table}

\paragraph{Complete GPU update.}
The FA3 benchmark measures generation, commitment and TAEHV decoding jointly (Table~\ref{tab:unfused_budget}). Deployment options are per-latent modulation, fused QKV projection, a fused GELU epilogue and fused commitment. Both rows retain the shared KV ring, FA3 and CUDA graphs. Both configurations complete all 100 updates within the 250\,ms budget. Host preparation, scheduling and display waits are outside this GPU endpoint. The FA2 and FA3 deployment environments also differ in PyTorch version, so each table isolates its own comparison.

\begin{table}[!tb]
\caption{\textbf{Eight-future GPU budget with FA3.} Times include generation, commitment and TAEHV decoding (ms); static compilation, one exclusive H100.}
\centering\small
\setlength{\tabcolsep}{3pt}
\begin{tabular}{lrrr}
\toprule
FA3 total update & P50 & P90 & P99\\
\midrule
Deployment options off & 235.1 & 236.8 & 238.0\\
All options on & 229.9 & 231.2 & 232.8\\
\bottomrule
\end{tabular}

\label{tab:unfused_budget}
\end{table}

Fused commitment moves the clean-history update into generation. With other options enabled, total GPU time is 228.6\,ms with separate commitment and 229.9\,ms with fused commitment. The post-decision path shortens from approximately 17.5 to 3.5\,ms, including decoding. This relocation makes a prepared latent available sooner after input.

\paragraph{Playback-time capacity.}
In the main eight-action, one-frame-threshold runs, Reactive/One Future/Eight Futures use 19\%/25\%/80\% GPU busy time and achieve 15.75/15.79/15.99\,fps. Extra preparation occupies computation available while committed frames play. The next section reports both commit and display endpoints on these same traces.

\section{Continuous-Stream Response}
\label{sec:stream_details}
Tables~\ref{tab:commit_decomposition}--\ref{tab:equal_rate_effects} expand Tables~\ref{tab:system_response} and~\ref{tab:main_latency}. The six-seed runs share their scoring window and equal-rate estimator (Sec.~\ref{sec:supp_protocol}); they separate deployment, queue threshold and scheduling policy.

\begin{table*}[!tb]
\caption{\textbf{Commit and presentation endpoints.} Six seeds, same 1857 scored inputs, equal-rate-mixture all-press quantiles in ms; unobserved successors remain infinite.}
\centering\small
\begin{tabular}{llcc}
\toprule
Deployment & Read-ahead & Input-to-commit P50/P90/P99 & Input-to-display P50/P90/P99\\
\midrule
SolarWM-FT B3, FastTree & Auto (4 frames) & 453/823/1336 & 620/983/1385\\
SolarWM-FT B3, FastTree & 1 frame & 493/871/1313 & 536/915/1370\\
ShowNow B1, FastTree & Auto (3 frames) & 169/298/328 & 340/468/499\\
ShowNow B1, FastTree & 1 frame & 163/292/327 & 210/339/374\\
ShowNow Reactive, optimized & 1 frame & 136/269/298 & 211/344/373\\
ShowNow One Future, optimized & 1 frame & 86/221/258 & 211/344/374\\
ShowNow Eight Futures, optimized & 1 frame & 88/222/250 & 208/345/374\\
\bottomrule
\end{tabular}

\label{tab:commit_decomposition}
\end{table*}

The FastTree one-frame comparison gives B3/B1 display medians of 536/210\,ms. B3 achieves 14.50\,fps with 5.91\% ongoing empty ticks; four-frame automatic lead restores 15.76\,fps with a 620\,ms median. Within B1, optimized execution reduces busy time from 28\% to 19\% and input-to-commit median by 27\,ms (95\% CI [18,33]). The main table reports the full deployment contrast.

\begin{table*}[!tb]
\caption{\textbf{Response by input rate.} Input-to-display P50/P90/P99 (ms) from the main six-seed runs. The mixture column reproduces the equal-rate estimator; each remaining column scores one input rate on the same traces.}
\centering\small\setlength{\tabcolsep}{4pt}
\begin{tabular}{lccccc}
\toprule
Deployment & Equal-rate mixture & 0.5 events/s & 1 event/s & 2 events/s & 4 events/s\\
\midrule
ShowNow, Reactive & 211/344/373 & 207/334/371 & 224/346/371 & 199/342/374 & 215/349/498\\
ShowNow, One Future & 211/344/374 & 211/343/373 & 213/341/370 & 207/347/374 & 213/347/475\\
ShowNow, Eight Futures & 208/345/374 & 226/338/371 & 207/352/371 & 209/355/521 & 197/331/453\\
SolarWM-FT, automatic lead & 620/983/1385 & 587/934/1035 & 613/1012/2423 & 627/1014/1602 & 654/969/1087\\
\bottomrule
\end{tabular}
\label{tab:rate_detail}
\end{table*}
\paragraph{Where time is spent.}
Table~\ref{tab:press_breakdown} decomposes directly served inputs into decision wait, post-decision generation, decoding and queue wait. Component means weight rates equally within the directly served subsets; all-press quantiles also include successor waits. One Future can regenerate before the logged selection decision, so its zero post-decision generation entry captures only the remaining work.

\begin{table*}[!tb]
\caption{\textbf{Directly served input decomposition at 16 fps.} Entries are equal-rate-weighted means in ms over directly served inputs in the first 50 s of the six-seed traces. Components sum per event; rounded entries can differ by 1 ms. The all-press quantiles additionally include successor waits. Generation is measured after the selection decision.}
\centering\small
\begin{tabular}{lrrrrrr}
\toprule
Deployment & Served inputs & Decision wait & Generation & Decode & Queue wait & Presentation\\
\midrule
ShowNow Reactive & 1730 & 92 & 49 & 3 & 63 & 208 \\
ShowNow One Future & 1732 & 104 & 0 & 4 & 100 & 207 \\
ShowNow Eight Futures & 1741 & 93 & 0 & 3 & 99 & 196 \\
B1 FastTree Reactive, threshold 1 & 1725 & 89 & 78 & 3 & 41 & 211 \\
SolarWM-FT, automatic threshold & 1032 & 218 & 144 & 10 & 135 & 506 \\
\bottomrule
\end{tabular}

\label{tab:press_breakdown}
\end{table*}

Automatic-lead SolarWM-FT serves 1032 inputs directly, 595 through a successor without idle and 230 across idle. Eight Futures gives 1741/89/27; both total 1857. SolarWM-FT minus Eight Futures display differences are 412/638/1011\,ms, with paired 95\% intervals [392,425]/[622,650]/[687,2051].

\paragraph{Policy and queue effects.}
At one-frame threshold, Eight minus One Future display differences are $-3/+1/0$\,ms. At zero-frame threshold, Eight minus Reactive differences are $-8/-2/-79$\,ms. Table~\ref{tab:equal_rate_effects} gives their intervals. The accelerated 32\,fps runs use the same model restricted to four actions, playing 16\,fps-trained content at $2\times$ speed. At zero threshold, Four minus One Future differences are $-5/-4/-2$\,ms. Main-paper Table~3 reports all policy--threshold operating points, including FPS and both empty-tick rates.

\begin{table*}[!tb]
\caption{\textbf{Six-seed paired presentation differences (ms).} Equal-rate-mixture quantiles, 2000 paired seed-cluster bootstrap draws. Negative values favor multiple futures. The 32\,fps result uses accelerated playback and zero-frame threshold. }
\centering\small
\setlength{\tabcolsep}{4pt}
\begin{tabular*}{\textwidth}{@{\extracolsep{\fill}}lccc@{}}
\toprule
Setting / contrast & $\Delta$P50 [95\% CI] & $\Delta$P90 [95\% CI] & $\Delta$P99 [95\% CI]\\
\midrule
16 fps, lead 1: Eight minus One Future & -3 [-8,+4] & +1 [-9,+7] & +0 [-1,+94]\\
16 fps, lead 0: Eight minus Reactive & -8 [-22,+7] & -2 [-8,+6] & -79 [-154,+35]\\
32 fps, lead 0: Four minus One Future & -5 [-8,+0] & -4 [-6,-1] & -2 [-3,-1]\\
\bottomrule
\end{tabular*}

\label{tab:equal_rate_effects}
\end{table*}

\section{Controlled Action Changes}
\label{sec:supp_response_window}
\paragraph{Protocol and endpoints.}
The controlled study comprises 2650 matched pairs (5300 four-second model traces) on six H100s on a shared node: 600 eight-action lead-sweep pairs, 450 four-action lead-sweep pairs, 1000 uniform-phase pairs and 600 brief-input burst pairs. The evaluation uses the eight-action, three-step student, 15 real history latents, optimized BF16/FA3 CUDA graphs and TAEHV. Four-action runs restrict the same model to WASD. Method order is counterbalanced and configuration order reversed on alternate GPUs; each pair shares a scene, input and device. Ten scenes, all trial counts and input seeds are fixed before measurement. Both policies share a 6\,ms selection guard and four-frame FIFO. The final changed command arrives $L$ before a 2\,s target and remains held through the end of the trace. We count a target hit when its first selected frame is scheduled no later than target$+2$\,ms. Software presentation includes generation, commitment, decoding and queues. Both policies use the same input availability and commit rules. Every final command is served; unserved commands are assigned infinite latency.

The eight-action sweep tests leads of 0, 2, 4, 6, 8, 10, 15, 20, 30, 40, 50, 60, 80, 100, 125, 150, 175, 200, 225 and 250\,ms, with 30 pairs per point. The four-action sweep uses the same grid through 125\,ms. Separate controls sample $L$ uniformly over the complete period, with 500 pairs per interface. Paired bootstrap intervals use 5000 resamples stratified by GPU; scene-cluster intervals provide a sensitivity check. Pointwise target-hit intervals are Wilson intervals.

\begin{table*}[!tb]
\caption{\textbf{Uniform-phase response controls.} Each interface uses 500 paired GPU switches. Input lead is uniform over the full four-frame period: 250\,ms at 16\,fps and 125\,ms at 32\,fps. The latter plays 16\,fps-trained content at $2\times$ speed.}
\centering\small
\begin{tabular}{llcc}
\toprule
Interface & Preparation & Display P50/P90/P99 (ms) & Target met (\%)\\
\midrule
Eight actions, 16 fps & One Future & 139.5/236.6/253.8 & 83.8\\
 & Eight Futures & 139.3/235.4/253.2 & 96.8\\
Four actions, 32 fps & One Future & 71.1/115.8/129.5 & 64.0\\
 & Four Futures & 67.0/117.2/130.7 & 89.0\\
\bottomrule
\end{tabular}
\label{tab:uniform_phase}
\end{table*}
For eight actions, the paired mean latency reduction is 9.25\,ms (95\% CI [7.00,12.00]; scene-cluster CI [6.75,11.75]), and target attainment increases by 13.0 percentage points [10.2,16.0]. Four actions give 9.19\,ms [7.44,10.94] and 25.0 points [20.6,29.6]. The full-distribution percentiles in Table~\ref{tab:uniform_phase} remain close; the improvement is concentrated in meeting the display deadline.

\begin{figure*}[!tb]
\centering
\includegraphics[width=\textwidth]{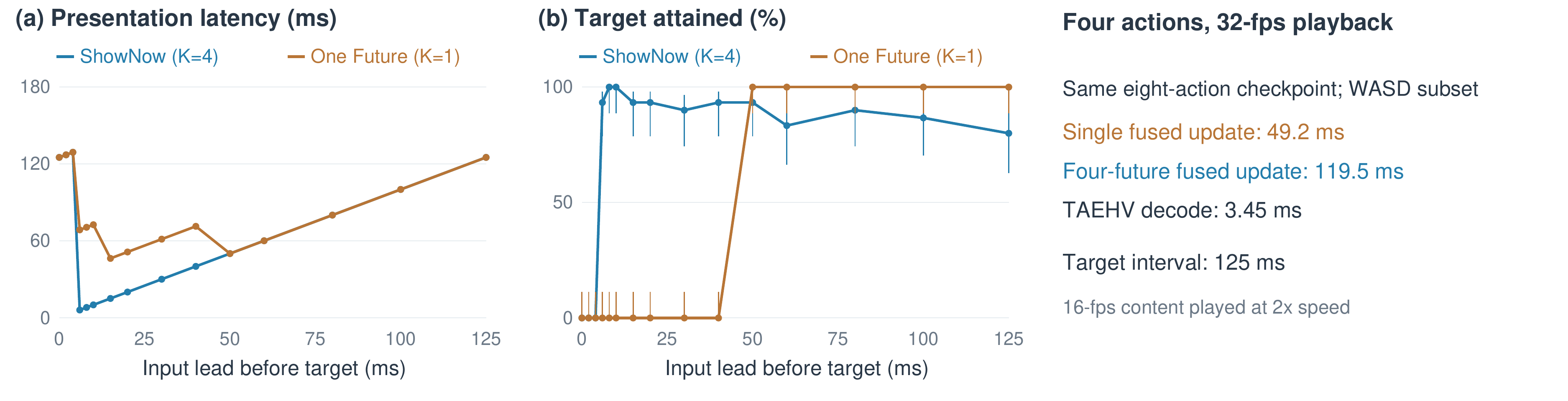}

\vspace{-2.5mm}

\caption{\textbf{Measured four-action response window at a 32\,fps playback target.} Thirty paired trials per lead, with the same checkpoint and scheduling rules as Figure~\ref{fig:response_compute}. Error bars show median-bootstrap or Wilson 95\% intervals. The lead sweep measures target attainment across input phases; beyond 50\,ms lead, single-future preparation often meets the target too. Playback is $2\times$ the 16\,fps training rate.}
\label{fig:supp_four_window}
\end{figure*}

Figure~\ref{fig:supp_four_window} extends the eight-action lead sweep in the main text to the four-action, 32\,fps interface. With stationary arrivals and independent display phase, lead to a periodic target is uniform over the period. The 6--40\,ms window therefore occupies 13.6\% of a 250\,ms period and 27.2\% of a 125\,ms period. The uniform-phase control measures the average over the complete period. At leads of 50--125\,ms, Four Futures has some lower target-hit estimates than One Future. Its 119.5\,ms fused update plus 3.45\,ms decode leaves approximately 2\,ms within the 125\,ms period; the sweep retains candidate-readiness and scheduling misses.

\paragraph{Brief-input bursts.}
The 600 burst pairs cover both interfaces at mean inter-command gaps of 30, 50 and 80\,ms, with 100 pairs per setting. Each burst has five commands; the final command arrives 10--40\,ms before the target and remains held. Eight Futures meets that target in all trials, versus 0--8\% for One Future; the four-action rates are 92--99\% versus 0--2\%. At 16\,fps and 30/50\,ms gaps, neither policy displays any of the 400 intermediate commands before their successors. These trials measure final-command settling; the uniform-phase controls in Table~\ref{tab:uniform_phase} cover the complete arrival period.

\section{Generation Cost at Native Playback}
\label{sec:native_cost_sweep}
We run the student on an exclusive H100 with optimized execution, four actions (W/S/A/D), one-frame queue threshold and native 16\,fps playback. Each policy receives identical traces at the four input rates in Sec.~\ref{sec:supp_protocol}, 60\,s per trace and three seeds (921 scored inputs per policy per sampling-step setting). Increasing sampling steps from three to four to six raises single-candidate generation plus fused commitment from 50 to 63 to 89\,ms; four-candidate updates cost 121/156/226\,ms. Those update costs include generation and fused commitment. The presentation measurements below additionally include TAEHV decoding and queueing.

\begin{table*}[!tb]
\caption{\textbf{Measured native-playback cost sweep.} Presentation quantiles in ms from equal-rate mixtures over 921 scored inputs (first 50 s of 60 s traces). Superseded inputs retain successor waiting. All configurations have zero ongoing empty ticks. Four Futures achieves 16.00\,fps at six steps, versus 15.56/15.64 for Reactive/One Future.}
\centering\small
\begin{tabular}{clccccc}
\toprule
Steps & ShowNow policy & P50 & P90 & P99 & Restart empty ticks (\%) & GPU busy (\%)\\
\midrule
3 & Reactive & 219 & 345 & 373 & 1.56 & 19\\
3 & One Future & 216 & 348 & 373 & 1.24 & 25\\
3 & Four Futures & 226 & 349 & 373 & 0.05 & 43\\
\midrule
4 & Reactive & 215 & 345 & 374 & 1.78 & 23\\
4 & One Future & 216 & 346 & 373 & 1.54 & 31\\
4 & Four Futures & 223 & 349 & 373 & 0.06 & 56\\
\midrule
6 & Reactive & 217 & 348 & 498 & 2.77 & 32\\
6 & One Future & 215 & 344 & 374 & 2.24 & 43\\
6 & Four Futures & 211 & 345 & 373 & 0.03 & 81\\
\bottomrule
\end{tabular}

\label{tab:native_cost_sweep}
\end{table*}

With increasing sampling cost, Four Futures' paired commit-median advantage over Reactive grows from 43 to 54 to 84\,ms (95\% CIs for Four minus Reactive: $[-48,-37]$, $[-57,-48]$, $[-107,-74]$). Presentation P50 differences are $+7\,[2,13]$, $+8\,[4,14]$ and $-6\,[-25,4]$\,ms. Broader preparation incurs a small median cost at three and four steps while improving restart continuity. At six steps, its presentation P99 difference is $-125\,[-189,-103]$\,ms; One Future minus Reactive also gives a P99 difference of $-124\,[-188,-100]$\,ms. Bootstrap estimates use 2000 paired resamples of three seed clusters. Differences are computed before rounding displayed quantiles.

\section{Control and Generation Quality}
\label{sec:supp_test}
\paragraph{Independent states and conditioning.}
Table~\ref{tab:e2_extended} reports the fixed eight-action student on validation and independent test states. Normal inference recomputes history with a cache-free tree sampler; real-time inference caches history and prepares eight candidates. Conditioning controls remove actions and/or camera motion at inference without retraining. Removing the action input at inference reduces validation accuracy from .874 to .200; additionally freezing the future camera gives .139. The independent test uses 96 states disjoint from the validation states used for model selection.
\begin{table*}[!tb]
\caption{\textbf{E2 quality and conditioning.} Validation/test use 48/96 states with eight futures each, the frozen switch-aware IDM and Wan VAE. LPIPS and PSNR cover the first second. Input-removal controls are evaluated on validation states.}
\centering\small
\begin{tabular}{lccc|ccc}
\toprule
 & \multicolumn{3}{c}{Validation} & \multicolumn{3}{c}{Independent test}\\
Inference / condition & Accuracy$\uparrow$ & LPIPS$\downarrow$ & PSNR$\uparrow$ & Accuracy$\uparrow$ & LPIPS$\downarrow$ & PSNR$\uparrow$\\
\midrule
SolarWM-FT & .793 & .401 & 18.80 & .791 & .394 & 18.96\\
ShowNow, normal & .874 & .416 & 18.06 & .852 & .415 & 18.02\\
ShowNow, real time & .860 & .437 & 17.64 & .853 & .427 & 17.87\\
Normal, no action, static camera & .139 & .494 & 16.83 & --- & --- & ---\\
Normal, camera only & .200 & .495 & 16.68 & --- & --- & ---\\
\bottomrule
\end{tabular}
\label{tab:e2_extended}
\end{table*}
\paragraph{Runtime decoder.}
Using TAEHV on the same saved E2 validation rollouts, while keeping the Wan-decoded real reference, gives one-second LPIPS/PSNR/SSIM of .407/19.63/.572 for the teacher, .422/18.89/.559 for SolarWM-FT and .456/17.76/.530 for ShowNow. TAEHV increases LPIPS by .019--.034 across models and horizons; PSNR changes by at most .13\,dB and SSIM by at most .002. Model ordering is preserved. Real latents decoded by TAEHV versus Wan VAE give LPIPS .141, PSNR 30.39\,dB and SSIM .834. Latent IDM scores are unchanged.

\paragraph{Feature distance.}
For each latent $z\in\mathbb R^{48\times H\times W}$, concatenate its per-channel spatial means and standard deviations into $\phi(z)\in\mathbb R^{96}$. Each five-second generated window contributes 20 latents per scene, pooled across 20 scenes. The fixed real pool contains all stored latents from the eight recorded branches of those scenes and is shared across windows. With empirical means and sample covariances, we compute
\begin{equation}
\begin{split}
\mathrm{FD}={}&\|\mu_g-\mu_r\|_2^2+\mathrm{tr}(\Sigma_g+\Sigma_r)\\
&-2\sum_i\sqrt{\max(\mathrm{Re}\,\lambda_i(\Sigma_g\Sigma_r),0)},
\end{split}
\end{equation}
where $\lambda_i$ are eigenvalues. Features use float-converted latents without further normalization. Minute-average FD averages twelve window scores, then noise seeds; endpoint FD uses the final five-second window at the specified update count.

\paragraph{One-minute rollouts.}
Table~\ref{tab:e3_three_seed} expands the long-horizon results in the main text on 20 held-out scenes. Deployed and fusions-off configurations use nine paired noise seeds; normal inference and SolarWM-FT use three seeds. Fusions-off disables the four options listed in Sec.~\ref{sec:supp_compute}, retaining the KV ring and FA3. Quality rollouts execute with explicit noise tensors; runtime benchmarks use CUDA graphs. FD intervals use 1000 two-level scene-and-seed percentile bootstrap draws, rebuilding feature pools after resampling scenes. Scene draws are shared across configurations; deployed/fusions-off seeds are paired and reference seeds are resampled independently.

\begin{table*}[!tb]
\caption{\textbf{One-minute E3 quality across noise seeds.} Same 20 test scenes, keys and nominal cameras. FD pools scenes within each five-second window and averages twelve windows. Ranges span noise seeds; intervals use the two-level bootstrap described in the text.}
\centering\small
\begin{tabular}{lcccc}
\toprule
E3 configuration & Noise seeds & Mean FD (seed range) & FD 95\% CI & Mean key accuracy (seed range)\\
\midrule
ShowNow, real time & 9 & 11.4 (6.2--17.3) & [6.16, 22.83] & .772 (.735--.794)\\
ShowNow, fusions off & 9 & 8.3 (5.5--11.4) & [5.21, 16.10] & .785 (.766--.807)\\
ShowNow, normal inference & 3 & 6.3 (4.3--9.8) & [4.21, 13.52] & .801 (.790--.813)\\
SolarWM-FT & 3 & 3.4 (3.3--3.5) & [3.25, 5.10] & .912 (.904--.922)\\
\bottomrule
\end{tabular}

\label{tab:e3_three_seed}
\end{table*}

Relative to deployment, fusions-off, normal inference and SolarWM-FT have FD differences of $-3.06$, $-5.12$ and $-7.94$, with two-level 95\% intervals $[-11.08,1.70]$, $[-14.89,1.18]$ and $[-18.30,-2.38]$. The corresponding nine-seed fusions-off key-accuracy difference is $+.014$ (two-level interval $[-.012,.047]$). Short-horizon deployed/fusions-off accuracy is .860/.854 on validation and .853/.851 on test, with test LPIPS .427/.427. Long-horizon fidelity varies across inference paths and rollout depths; the intervals report the observed seed and scene variation.

\paragraph{Aligned autoregressive depth.}
B1 performs 240 updates per minute; B3 performs 80. At 80 updates, three-seed endpoint FD is 4.0 for normal B1, 9.8 for deployed B1 and 4.1 for SolarWM-FT. These endpoints represent 20\,s for B1 and 60\,s for B3, using five-second endpoint windows across the same 20 scenes. Their scripted histories contain approximately ten/thirty switches. This diagnostic aligns update count while duration and encountered actions differ.

\subsection{Detected Action-Effect Latency}
\label{sec:action_effect}
We additionally evaluate the same 96 held-out E2 test states used in the independent quality evaluation, disjoint from the main-table validation scenes. Each 8\,s episode starts with a press at 0.5\,s, followed by isolated changes every 1.0--1.25\,s; only presses through 6\,s are scored, giving 500 events per configuration. ShowNow runs at 16\,fps with eight actions and one-frame lead; the three policies are interleaved by scene, with reversed order for the second half. SolarWM-FT uses trained B3 with automatic lead. Both columns in Table~\ref{tab:action_effect} use the same 500 events.
\begin{table}[!tb]
\caption{\textbf{Offline action-response proxy on isolated switches.} P50/P90/P95 in ms. The IDM endpoint assigns recognition to the central latent and retains successor waiting; display uses the same 500 inputs.}
\centering\small
\begin{tabular}{lcc}
\toprule
Configuration & IDM / successor & Display\\
\midrule
ShowNow, Reactive & 297/711/1061 & 229/348/361\\
ShowNow, One Future & 302/700/1072 & 230/349/362\\
ShowNow, Eight Futures & 297/815/1238 & 230/348/361\\
SolarWM-FT & 603/987/1027 & 603/962/1008\\
\bottomrule
\end{tabular}

\label{tab:action_effect}
\end{table}

SolarWM-FT minus Eight Futures has detected-effect differences of 306/173/$-210$\,ms, with paired scene-cluster 95\% intervals $[282,336]$/$[34,288]$/$[-377,142]$ (2000 resamples). Median and P90 favor ShowNow; the P95 point estimate favors SolarWM-FT and its interval crosses zero. 

Detection uses a frozen IDM over latents $(t-1,t,t+1)$ and assigns the result to the first displayed frame of latent $t$. Detection is quantized to one latent (250\,ms); the $t+1$ context can assign an effect up to one latent early. Successor timing measures the wait until a later command appears when the original command is superseded. Unobserved successors retain $+\infty$. The fractions without an observed effect or successor are 0.8\%, 0.4\%, 0.2\% and 0.0\% for Reactive, One Future, Eight Futures and SolarWM-FT, respectively. The endpoint therefore combines action detection with successor waiting.

\section{ShowNow-33B Extension}
\label{sec:h3_extension}
ShowNow-33B fine-tunes the released five-latent, four-step SolarWM-H3 generator for single-latent, three-step generation with an optimized engine. Generation and commitment are timed with eight-way B300 sequence parallelism at $1344\times768$. The evaluation models an additional, independent decoder GPU: decoding is represented by a fixed 47\,ms per latent, calibrated from a separate B300 SolarRenderer demo using the same H3 VAE and resolution. Generation and decoding overlap in this model.

A single-latent result becomes available at commitment time plus 47\,ms. For a five-latent block, latent $j$ becomes available at commitment time plus $(j+1)47$\,ms, for $j=0,\ldots,4$. A playback thread consumes available frames in order at 24\,fps and records empty ticks when the next frame is unavailable. Latency runs from input to scheduled presentation of the first frame conditioned on that input, retaining successor waiting for superseded commands. The protocol uses six seeds, four refractory-Poisson rates, 60\,s traces and 1862 scored inputs in the first 50\,s; quantiles use an equal-rate mixture. Inter-GPU transfer, client networking and physical display delay are excluded. 

\begin{table*}[!tb]
\caption{\textbf{ShowNow-33B hybrid timing study.} GPU generation and commitment are measured; an independent ninth GPU is modeled with fixed 47\,ms decoding per latent. Empty time is normalized by held-key time. The single-latent rows use the single-latent inference engine; the native model uses four sampling steps and the adapted model three. Working points vary selection lead.}
\centering\small
\begin{tabular}{lcrrr}
\toprule
Deployment & Selection lead & P50/P90/P99 (ms) & Empty time (\%) & Playback FPS\\
\midrule
SolarWM-H3, native Reactive & Automatic & 862/1138/1202 & 0.0 & 24.06\\
ShowNow-33B, Reactive & Automatic & 311/396/413 & 0.0 & 24.12\\
ShowNow-33B, One Future & 1 frame & 167/229/249 & 8.4 & 22.08\\
ShowNow-33B, One Future & 2 frames & 196/269/290 & 2.2 & 23.58\\
ShowNow-33B, One Future & 3 frames & 232/307/330 & 0.8 & 23.94\\
\bottomrule
\end{tabular}

\label{tab:h3_hybrid}
\end{table*}

The single-latent Reactive working point reduces median modeled-pipeline latency by 64\% relative to the native deployment. One Future with three-frame lead further lowers the median from 311 to 232\,ms, with 0.8\% empty time. Block size, sampling steps, runtime and selection policy contribute jointly to these deployment differences. A one-frame lead is 41.7\,ms, below the assumed decode time; two and three frames provide 83.3 and 125\,ms. The calibration demo measures about 47\,ms for the first latent and 49\,ms after startup; the timing model uses the fixed first-latent value throughout.

Quality evaluation on 32 videos reports PSNR gaps of 1.5--2.1\,dB relative to the native model. From static history, action directions are generally correct with weaker forward motion; conflicting commands can fail to override existing motion history. Latency tracks newly conditioned frames independently of whether the commanded motion is correctly expressed. Timing and the 32-video quality diagnostic use different adaptation checkpoints. Figure~\ref{fig:h3_showcase} is a separate recorded-camera showcase; these measurements and selected visuals describe their respective evaluations.

\section{Scope and Failure Modes}
\label{sec:supp_limitations}
\paragraph{Response and command retention.}
Response can exceed one 250\,ms update when a command is superseded before commitment: successor waiting can span later inputs or released-key intervals. Decoding and queueing add to presentation latency. All such events remain in the continuous-stream estimator. At rapid input rates, multiple changes can occur within one latent, so the committed stream follows the selected command and may omit intermediate commands.

\paragraph{Scene memory and fidelity.}
The framework optimizes response scheduling and supports short-horizon action changes. It uses bounded visual history, with no explicit persistent scene memory or revisit-consistency objective. Long rollouts can drift in appearance or reconstruct revisited regions differently. The multi-seed results in Sec.~\ref{sec:supp_test} quantify the accompanying quality changes. Shared attention is algebraically equivalent over the allowed partitions; cached and cache-free autoregressive executions can still accumulate numerical and history differences.

\paragraph{Data and action coverage.}
Matched training futures require a simulator or game engine that can restore a common state and record alternative outcomes and camera poses. Multi-action evidence covers eight mutually exclusive ViZDoom controls and synthetic input traces; the 33B extension evaluates single-future camera-conditioned generation across broader scenes. Larger discrete or continuous control spaces call for limited candidate budgets guided by action predictions or sampled proposals. Real-player traces and perceptual evaluation would connect scheduled-frame gains to user experience.

\section{Qualitative Examples}
\label{sec:supp_qualitative}
The following pages separate simulator observations, multi-future predictions and continuous generated streams. The ViZDoom panels emphasize readable scenes and retain their original brightness. The 33B showcase uses selected sequences across indoor, outdoor and game scenes; its source selection filters collapsed or severely blurred outputs and ranks LPIPS within each source.

\begin{figure*}[!t]
\centering
\includegraphics[width=\textwidth,keepaspectratio]{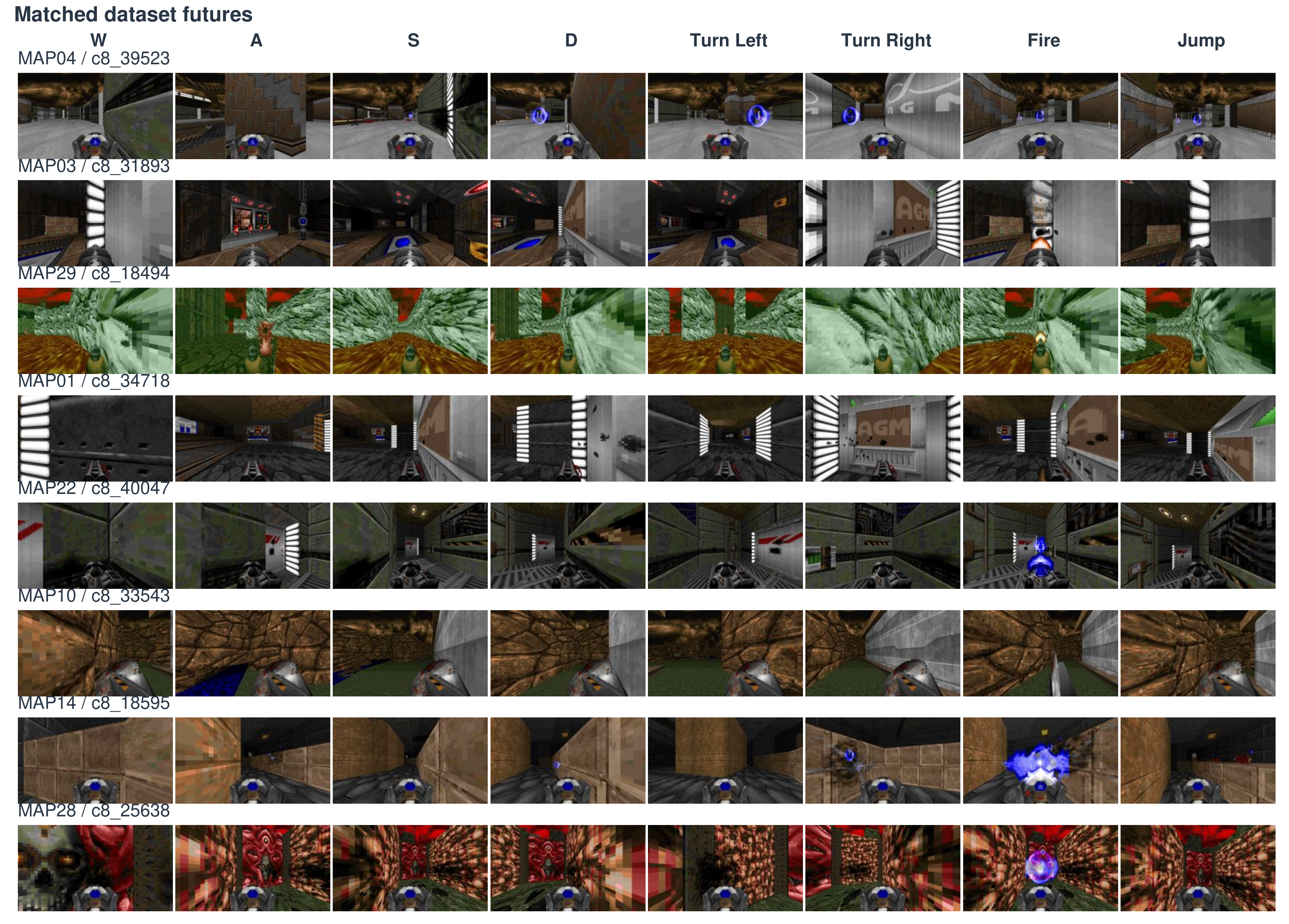}

\vspace{-2.5mm}

\caption{\textbf{Matched dataset futures.} Each row shares one simulator state; columns show eight fixed actions at 1.5\,s (frame 47 of the 32\,fps source). All panels are simulator observations.}
\label{fig:supp_dataset}
\end{figure*}

\begin{figure*}[!p]
\centering
\includegraphics[width=\textwidth,height=0.87\textheight,keepaspectratio]{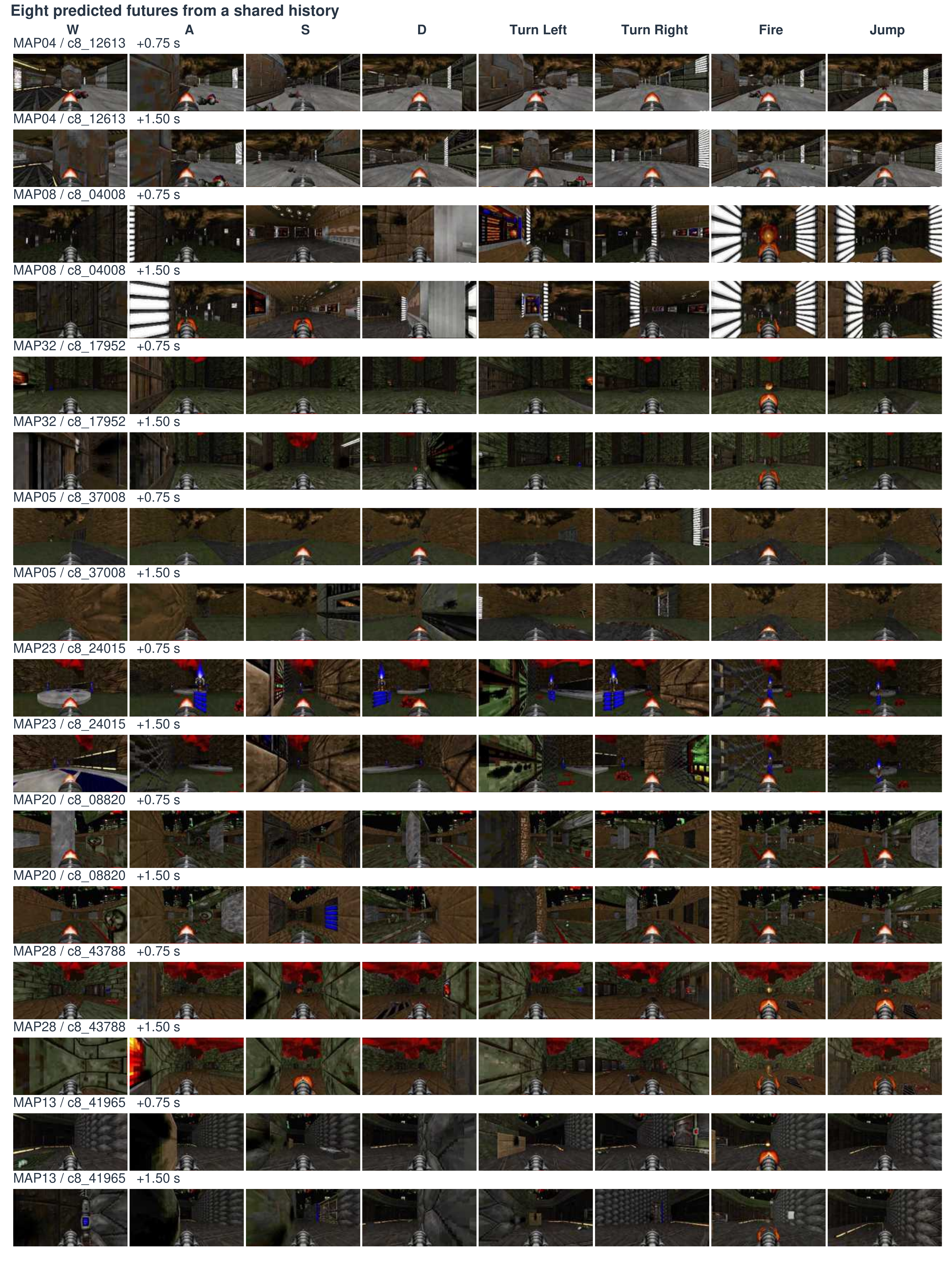}

\vspace{-2.5mm}

\caption{\textbf{Eight predictions from one history.} Each state appears at 0.75 and 1.5\,s; columns hold eight distinct actions throughout. Normal inference with Wan-VAE decoding; all panels are generated.}
\label{fig:supp_shared_futures}
\end{figure*}

\begin{figure*}[!p]
\centering
\includegraphics[width=\textwidth,height=0.87\textheight,keepaspectratio]{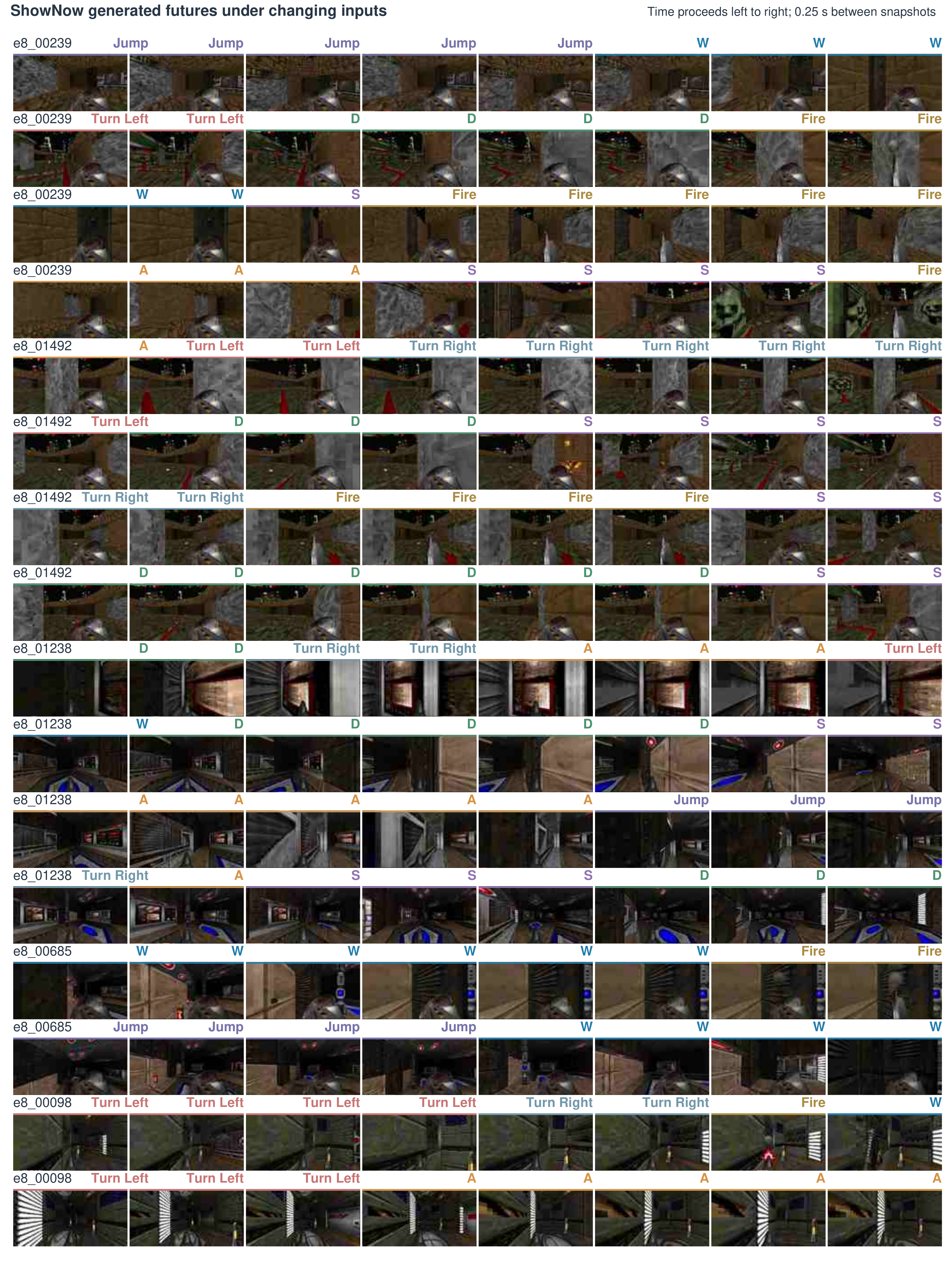}

\vspace{-2.5mm}

\caption{\textbf{Generated sequences under changing input.} Sixteen branches across five scenes from the real-time path, decoded with Wan VAE. Snapshots are 0.25\,s apart; labels show the action conditioning each displayed latent. These saved rollouts illustrate visual changes across inputs.}
\label{fig:supp_generated}
\end{figure*}

\begin{figure*}[!p]
\centering
\includegraphics[width=\textwidth,height=0.87\textheight,keepaspectratio]{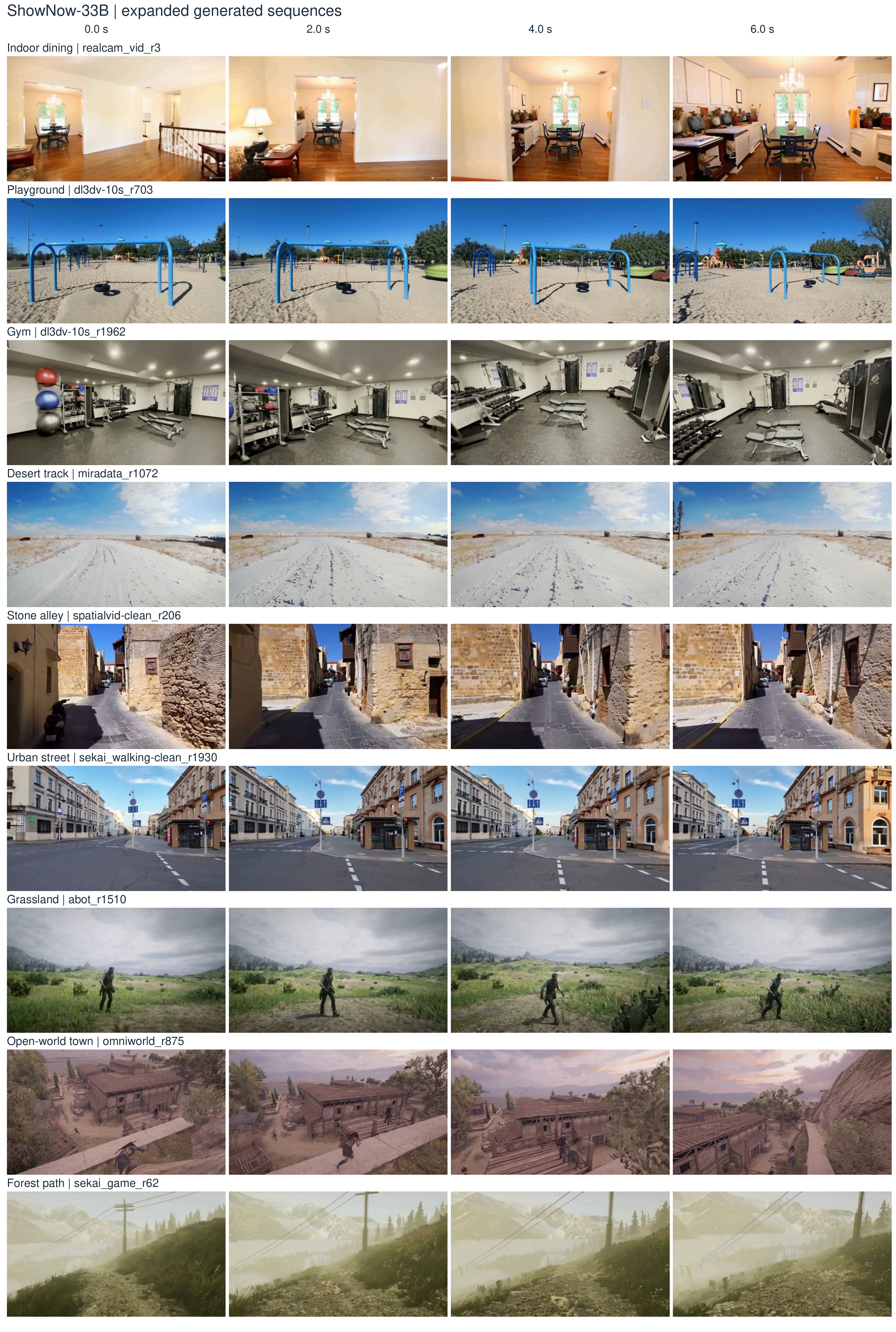}

\vspace{-2.5mm}

\caption{\textbf{ShowNow-33B expanded sequences.} The nine cases from Figure~\ref{fig:qualitative_results}(b), in row-major order, with scene descriptions and source IDs. Columns sample clip-relative times 0, 2, 4 and 6\,s under recorded camera trajectories. The single-latent 33B variant uses three sampling steps at $1344\times768$.}
\label{fig:h3_showcase}
\end{figure*}

Figure~\ref{fig:supp_dataset} shows eight shared simulator states under all eight actions. Figure~\ref{fig:supp_shared_futures} shows predictions at two times from common histories. Figure~\ref{fig:supp_generated} shows generated streams under changing keys. Figure~\ref{fig:h3_showcase} expands nine ShowNow-33B sequences at four timestamps, following recorded camera trajectories. The galleries illustrate matched data, action-conditioned futures, changing-input sequences and scene diversity. Quantitative timing is evaluated in Sections~\ref{sec:stream_details}--\ref{sec:native_cost_sweep}.

\clearpage

\clearpage

\end{document}